\documentclass[10pt,twocolumn,letterpaper]{article}

\usepackage[pagenumbers]{cvpr} 

\usepackage{times}
\usepackage{epsfig}
\usepackage{graphicx}
\usepackage{amsmath}
\usepackage{amssymb}
\usepackage{subcaption}

\usepackage{graphicx}
\usepackage{amsmath}
\usepackage{amssymb}
\usepackage{booktabs}
\usepackage{multicol}
\usepackage{multirow, makecell}
\usepackage{array}

\usepackage{float}

\usepackage[table]{xcolor}

\newcommand{\ccb}{\cellcolor{blue!10}}

\usepackage{enumitem}
\usepackage{color}

\usepackage[titletoc]{appendix}

\newcolumntype{x}[1]{>{\centering\arraybackslash}p{#1pt}}
\newcolumntype{y}[1]{>{\raggedright\arraybackslash}p{#1pt}}
\newcolumntype{z}[1]{>{\raggedleft\arraybackslash}p{#1pt}}
\newcolumntype{*}{>{\global\let\currentrowstyle\relax}}
\newcolumntype{^}{>{\currentrowstyle}}

\newlength\savewidth\newcommand\shline{\noalign{\global\savewidth\arrayrulewidth
  \global\arrayrulewidth 1pt}\hline\noalign{\global\arrayrulewidth\savewidth}}
\newcommand{\tablestyle}[2]{\setlength{\tabcolsep}{#1}\renewcommand{\arraystretch}{#2}\centering\footnotesize}
\definecolor{lightgreen}{HTML}{D8ECD1}
\newcommand{\better}[1]{\colorbox{lightgreen}{#1}}
\newcommand{\highlight}[1]{\colorbox{green}{#1}}

\definecolor{lavenderblue}{rgb}{0.8, 0.8, 1.0}
\definecolor{lavender(web)}{rgb}{0.9, 0.9, 0.98}

\usepackage[symbol*]{footmisc}

\newcommand\cellwidthbsample{1.3}

\newcommand\cellwidthsupp{1.3}

\newcommand\cellwidth{1.2}

\definecolor{cvprblue}{rgb}{0.21,0.49,0.74}
\usepackage[pagebackref,breaklinks,colorlinks,citecolor=cvprblue]{hyperref}

\def\paperID{2} 
\def\confName{CVPR}
\def\confYear{2024}

\title{Adapting Dual-encoder Vision-language Models for Paraphrased Retrieval}

\begin{document}

\author{
Jiacheng Cheng$^1$\footnotemark[1], Hijung Valentina Shin$^2$, Nuno Vasconcelos$^1$, Bryan Russell$^2$, Fabian Caba Heilbron$^2$
\\
$^1$UC San Diego \quad\quad\quad\quad $^2$Adobe Research
}

\twocolumn[{%
\maketitle

\newcounter{rownumber}[figure] 
\renewcommand{\therownumber}{1(\alph{rownumber})}

\newcommand\cellwidthteaser{1.4}
\begin{figure}[H]

\hsize=\textwidth 
\centering
\setlength{\tabcolsep}{1.5pt} 
\renewcommand{\arraystretch}{1.2}

\vspace{-0.9cm}
\begin{tabular}{ccccccccccc}
\multicolumn{11}{c}{{Query: \textit{“A young \textcolor{red}{kid} is holding a box of pizza.”}}} 
\tabularnewline
\includegraphics[width=\cellwidthteaser cm, height=\cellwidthteaser cm]{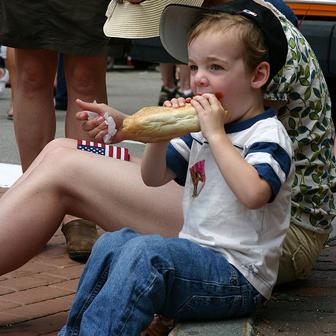}& 
\includegraphics[width=\cellwidthteaser cm, height=\cellwidthteaser cm]{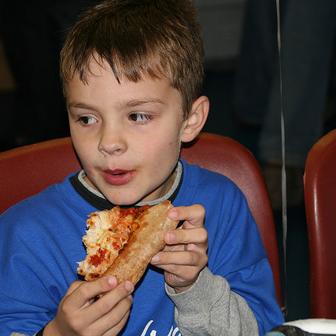}& 
\highlight{\includegraphics[width=\cellwidthteaser cm, height=\cellwidthteaser cm]{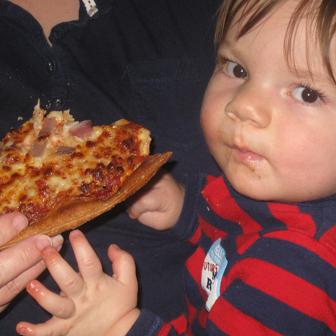}}& 
\includegraphics[width=\cellwidthteaser cm, height=\cellwidthteaser cm]{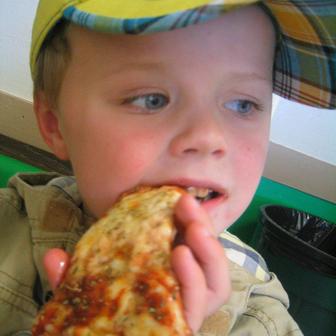}& 
\includegraphics[width=\cellwidthteaser cm, height=\cellwidthteaser cm]{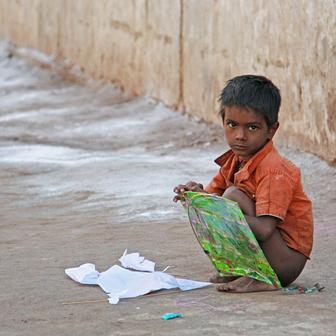}& 
\hspace{1cm} &
\highlight{\includegraphics[width=\cellwidthteaser cm, height=\cellwidthteaser cm]{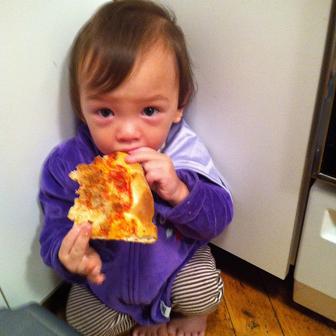}}& 
\highlight{\includegraphics[width=\cellwidthteaser cm, height=\cellwidthteaser cm]{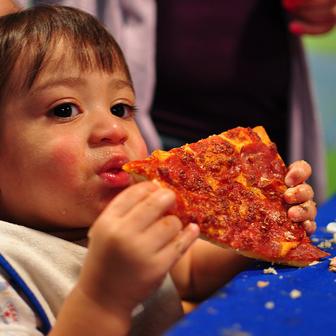}}& 
\highlight{\includegraphics[width=\cellwidthteaser cm, height=\cellwidthteaser cm]{samples/resized/teaser/000000567217.jpg}}&
\highlight{\includegraphics[width=\cellwidthteaser cm, height=\cellwidthteaser cm]{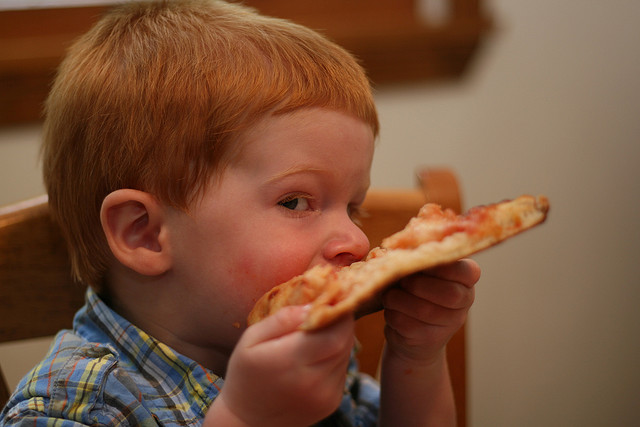}}&
\highlight{\includegraphics[width=\cellwidthteaser cm, height=\cellwidthteaser cm]{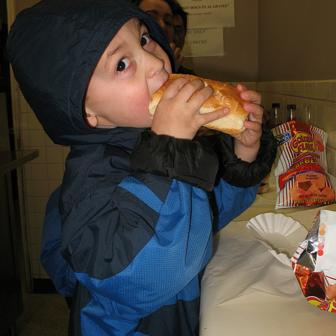}} 
\tabularnewline
\multicolumn{11}{c}{{Paraphrased query: \textit{“A young \textcolor{red}{child} is holding a box of pizza.”}}} 
\tabularnewline
\highlight{\includegraphics[width=\cellwidthteaser cm, height=\cellwidthteaser cm]{samples/resized/teaser/000000301951.jpg}}& 
\includegraphics[width=\cellwidthteaser cm, height=\cellwidthteaser cm]{samples/resized/teaser/000000196848.jpg}& 
\includegraphics[width=\cellwidthteaser cm, height=\cellwidthteaser cm]{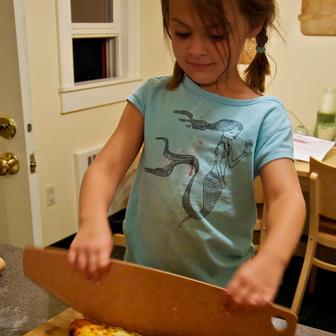}& 
\includegraphics[width=\cellwidthteaser cm, height=\cellwidthteaser cm]{samples/resized/teaser/000000111839.jpg}& 
\includegraphics[width=\cellwidthteaser cm, height=\cellwidthteaser cm]{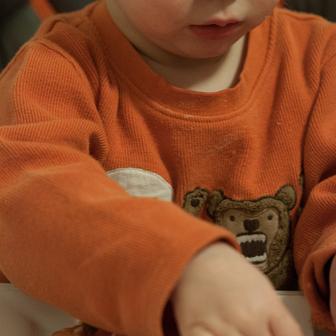}& 
~ &
\highlight{\includegraphics[width=\cellwidthteaser cm, height=\cellwidthteaser cm]{samples/resized/teaser/000000196848.jpg}}& 
\highlight{\includegraphics[width=\cellwidthteaser cm, height=\cellwidthteaser cm]{samples/resized/teaser/000000048395.jpg}}& 
\highlight{\includegraphics[width=\cellwidthteaser cm, height=\cellwidthteaser cm]{samples/resized/teaser/000000515787.jpg}}& 
\highlight{\includegraphics[width=\cellwidthteaser cm, height=\cellwidthteaser cm]{samples/resized/teaser/000000567217.jpg}}& 
\highlight{\includegraphics[width=\cellwidthteaser cm, height=\cellwidthteaser cm]{samples/resized/teaser/000000254319.jpg}}

\tabularnewline
\multicolumn{5}{c}{\refstepcounter{rownumber}\label{fig:teaser:left} {(a)} CLIP's top retrievals} & ~ &  \multicolumn{5}{c}{{(b)} \refstepcounter{rownumber}\label{fig:teaser:right} Our top retrievals}
\end{tabular}


\caption{\textbf{Paraphrased retrieval.} \ref{fig:teaser:left}: The CLIP model~\cite{radford2021learning} often returns very different top retrievals for two text queries that are paraphrases of each other. In this example, the paraphrased query differs by a single word: \textit{``kid''}$\rightarrow$\textit{``child''}. \ref{fig:teaser:right}: Our approach returns similar top retrievals for paraphrased queries by adapting a frozen pretrained language model during dual encoder training. We outline images in green that appear in the top retrievals for the original and paraphrased queries.}
\label{fig:teaser}

\end{figure}
}]

\footnotetext[1]{Work partially done during an internship at Adobe Research.}



\begin{abstract}
In the recent years, the dual-encoder vision-language models (\eg CLIP) have achieved remarkable text-to-image retrieval performance. However, we discover that these models usually results in very different retrievals for a pair of paraphrased queries. Such behavior might render the retrieval system less predictable and lead to user frustration. In this work, we consider the task of paraphrased text-to-image retrieval where a model aims to return similar results given a pair of paraphrased queries. To start with, we collect a dataset of paraphrased image descriptions to facilitate quantitative evaluation for this task. We then hypothesize that the undesired behavior of existing dual-encoder model is due to their text towers which are trained on image-sentence pairs and lack the ability to capture the semantic similarity between paraphrased queries. To improve on this, we investigate multiple strategies for training a dual-encoder model starting from a language model pretrained on a large text corpus. Compared to public dual-encoder models such as CLIP and OpenCLIP, the model trained with our best adaptation strategy achieves a significantly higher ranking similarity for paraphrased queries while maintaining similar zero-shot classification and retrieval accuracy. 
\end{abstract}

\section{Introduction}
\label{sec:intro}

The dual-encoder vision-language foundation models (VLFMs) such as CLIP~\cite{radford2021learning} have demonstrated promising zero-shot accuracy in text-image retrieval. However, in practice, their returned top retrievals are often very different for paraphrased queries. Consider the input text queries and the output visual search results produced by CLIP in Figure~\ref{fig:teaser:left}. 
In this example, when a single word is changed in the query (\textit{``kid''}$\rightarrow$\textit{``child''}), even though the semantic meaning of the query remains almost exactly the same, all but one of the images returned by CLIP are completely different. If small changes in the user prompt lead to very different output, the results can be interpreted as unpredictable. Consequently, the construction of appropriate queries can be challenging and lead to a negative user experience when the model is used in downstream search applications.

We conjecture that these results are caused by the training of the CLIP model starting from random initialization on a limited text dataset of 400M sentences (paired with images). This text dataset has the following shortcomings that impair effective learning of paraphrases. First, it is significantly smaller than datasets used for training large language models. Recent large language models are trained on datasets that are orders of magnitude larger~\cite{brown2020language}, allowing for greater exposure to paraphrase examples.
Second, the sentences are consumed in isolation and not in the context of other sentences and paragraphs. This latter context is essential for long-range reasoning in learning to paraphrase. Third, the alt texts used for training CLIP are often noisy and not even complete sentences~\cite{desai2021redcaps,entezari2023transfer,saharia2022imagen}, which may hinder paraphrase learning. 

Large language models pretrained on large text corpora have largely addressed these limitations. Moreover, these models have demonstrated impressive results on a variety of downstream language understanding tasks, including sentence semantic similarity~\cite{cer-etal-2017-semeval} and paraphrase detection ~\cite{dolan2005automatically}. There have been prior efforts attempting to guide vision-language model training with pretrained language models~\cite{li2023blip,alayrac2022flamingo,yang2022zeroshot}.  However, most of them are successful for fusion-encoder models (\eg BLIP-2~\cite{li2023blip}, Flamingo~\cite{alayrac2022flamingo}). 
LiT~\cite{zhai2022lit} proposed to adopt a frozen supervised pretrained model in the image tower and training from scratch or finetuning the text tower for best zero-shot classification accuracy. When they plugged in a frozen pretrained language model into the dual-encoder architecture, they showed a significant degradation in the zero-shot accuracy of the resulting model. Unfreezing and finetuning the pretrained text tower can improve the zero-shot retrieval/classification accuracy. However, as we will demonstrate in the experiments, this strategy suffers from vastly different retrieval results for paraphrased language queries. 
In this paper, we seek to address this key question: can we train a dual-encoder vision-language model guided by a strong language model pretrained on a large text corpora? An ideal model will maintain the prior zero-shot retrieval/classification accuracy while better handling paraphrased language queries. 

In this work, we show how to overcome the above limitations by guiding dual-encoder VLFMs training with pretrained language models. As our first contribution, we show that adding a set of alignment layers in the language tower significantly improves the zero-shot classification and retrieval accuracy compared to LiT's locked language tower setting~\cite{zhai2022lit}. 
As our second contribution, we collect a dataset of paraphrased sentences paired with natural images for evaluating whether paraphrased language queries result in different retrievals. We compute the rank similarity of paraphrased language queries and, as our final contribution, show empirically that our adaptation strategy results in a significant improvement, as illustrated in Figure~\ref{fig:teaser:right}.

The rest of this paper is structured as follows. The related work is discussed in Sec.\ \ref{sec:related}. In Sec.\ \ref{sec:dataset}, we introduce our collected dataset and evaluation metrics for quantitatively measuring the rank similarity in paraphrased text-image retrieval. Aiming to improve the rank similarity in paraphrased retrieval, we present multiple strategies for training dual-encoder models starting from a pretrained language model in Sec. \ref{sec:methods} and comprehensively compare them in Sec.\ \ref{sec:experiments}. In Sec.\ \ref{sec:conclusion}, we conclude our paper.

\section{Related Work}
\label{sec:related}

\noindent\textbf{Information Retrieval.}
\emph{Synonymy} refers to the problem that different query terms with same or similar meaning cannot be properly handled by a information retrieval system and finally lead to undesired outcomes (\eg low recall, user frustration). It is a well-known issue and has been extensively studied in the information retrieval literature~\cite{manning2008introduction,ModernInformationRetrieval}. To address it, many techniques have been proposed. Latent semantic indexing (LSI) \cite{deerwester1990indexing,landauer1998introduction} is able to discover relationships between words/terms in documents and thus reduce the effects of synonyms. However, the application of LSI for a dynamic corpus is prohibitive since it does not support ad-hoc addition of new documents \cite{gee2003using}. Two  popular alternative techniques are relevance feedback~\cite{rocchio71relevance,ruthven2003survey} and query expansion (QE)~\cite{carpineto2012survey,azad2019query}. Relevance feedback refines the search results according to the users' feedback on the initial search results, making it impractical when users are not involved in the retrieval process. QE expands the original query by adding semantically similar words from a pre-built thesaurus (\eg WordNet \cite{miller1995wordnet,miller1990introduction}). The performance of QE is usually sensitive to the quality of the word sense disambiguation in the query. In this work, we show that QE alone is not an effective solution to increase ranking similarity in paraphrased retrieval.

\noindent\textbf{Vision-Language Foundation Models (VLFMs).}
There are two mainstream categories for VLFMs: (1) Dual-encoder  models which have two separate image and text encoders \cite{radford2021learning,jia2021scaling,zhai2022lit,yao2022filip}. Dual-encoder models are usually pretrained by contrastive image-text learning and shown to be a strong zero-shot learner on many downstream cross-modal tasks (\eg, image classification and image/text retrieval). (2) Fusion-encoder models with additional cross-modal encoders \cite{chen2020uniter,zhang2021vinvl,li2021align,li2022blip,singh2022flava,li2023blip}. In general, fusion-encoder models are pretrained with a combination of unimodal, multi-modal, cross-modal objectives. Although fusion-encoder models are superior in many vision-language tasks, they are usually significantly less efficient than dual-encoder models for large-scale image/text retrieval. In light of this, we focus on dual-encoder models for paraphrased retrieval in this work.

\noindent\textbf{Robustness to Text Perturbations.}
According to a recent investigation~\cite{qiu2022multimodal}, the retrieval performance of many vision-language models including CLIP is sensitive to many text perturbations such as synonym replacement and word swap. Although these text perturbations may not be equivalent to paraphrasing, they are widely existing in real-world scenarios. In the experiment section, our proposed training strategy is  evaluated on different text perturbations and shown to achieve higher robustness. 

\section{Paraphrased Text-Image Retrieval}
\label{sec:dataset}
In order to perform quantitative evaluation of different vision-language models in paraphrased text-image retrieval, it is necessary to establish a dataset consisting of semantically equivalent natural language queries (paraphrase pairs) paired with corresponding images. We build on the MS COCO Captions multimodal dataset \cite{lin2014microsoft}, which features a large quantity of images with high-quality captions. We used the captions of its 5K validation images as source data for collecting paraphrase pairs. Finally, 4155 paraphrase pairs are collected by automatic paraphrase generation and manual verification. To simulate image databases at different scales, we construct two image galleries. The smaller one consists of 5K images in the COCO validation set. The larger one is formulated by the combination of 5K validation and 123K unlabeled images in COCO. These two settings are denoted as COCO Paraphrased retrieval and COCO Paraphrased retrieval Extended (shorted as COCO-P  and COCO-PE in the following) respectively. We describe next how we collect the paraphrased sentences and how we measure the rank similarity between the retrieval results for a query sentence and its paraphrase.

\noindent\textbf{Paraphrase Collection.}\label{sec:paraphrase_collect}
It is expensive and time consuming to generate high-quality sentence-level paraphrases at a large scale by human annotators. Consequently, traditional paraphrase datasets (\eg, Microsoft Research Paraphrase Corpus \cite{dolan2005automatically},  Quora Question Pairs \cite{QQP}) are constructed by first mining from newspapers/web sites followed by manual verification by a human. However, such approach does not serve our need since it is inefficient or even infeasible to find sentences with the equivalent meaning as certain image captions by web mining. 

An alternative approach is to generate paraphrase candidates for captions automatically. Automatic paraphrase generation is a longstanding and challenging task in natural language processing \cite{zhou2021paraphrase}. Fortunately, the success of large autoregressive language models with billions of parameters (\eg, GPT-2/3 \cite{radford2019language,brown2020language}) renders it possible to generate automatically high-quality paraphrase candidates in an unsupervised or few-shot manner \cite{witteveen2019paraphrasing, hegde2020unsupervised, Wahle2022d}. 

Inspired by these successful applications, we adopted GPT-3 for paraphrasing image captions. We chose 5000 captions describing the images in the validation set of the MS COCO Captions dataset as the source data. For each original caption, we generated 5 paraphrase candidates via 5 independent runs of GPT-3 with manually chosen hyper-parameters and prompts (see supplemental material). We show some examples in Fig.\ \ref{fig:para_samples}. In practice, we observe that most of the paraphrase candidates are of decent quality in terms of both correctness and diversity. To remove noisy paraphrase candidates that are incorrect or duplicates, we manually verified the paraphrase candidates with native English speaker human annotators. Following the annotation protocol in \cite{agirre-etal-2013-sem,cer-etal-2017-semeval}, the original and paraphrase candidate sentences are graded with a similarity score between 0 and 5. A sentence pair that are paraphrases and identical in meaning has a similarity score of 5 (\ie perfect paraphrase). 
After removal of duplicate and paraphrase pairs with similarity score less than 5, there are 4155 captions with at least one perfect paraphrase. For each of the captions, one perfect paraphrase was randomly chosen for our benchmark.


\begin{figure}[!t]
\centering
\footnotesize
\tabcolsep=0.02in
\begin{tabular}{|m{1.3cm}|l|}
\toprule
{\includegraphics[width=\cellwidthbsample cm, height=\cellwidthbsample cm]{samples/000000055150.jpg}} & \makecell[l]{\textbf{Paraphrasing Score}: 5 \\ 
\textbf{Query}:  \\
\textit{Two \textcolor{red}{children smiling on top of luggage} in parking lot}. \\ 
\textbf{Paraphrased query}: \\
\textit{Two \textcolor{red}{kids grinning atop suitcases} in a parking lot.}}
\tabularnewline
\midrule
{\includegraphics[width=\cellwidthbsample cm, height=\cellwidthbsample cm]{samples/000000206027.jpg}} &
\makecell[l]{\textbf{Paraphrasing Score}: 4 \\
\textbf{Query} : \\ 
\textit{A \textcolor{red}{slice} of pizza  that is on top of a \textcolor{red}{napkin}.} \\ 
\textbf{Paraphrased query}: \\ 
\textit{A \textcolor{red}{piece} of  pizza that is on top of a \textcolor{red}{paper towel}.}} 
\tabularnewline
\midrule
{\includegraphics[width=\cellwidthbsample cm, height=\cellwidthbsample cm]{samples/000000364297.jpg}} &
\makecell[l]{\textbf{Paraphrasing Score}: 3 \\
\textbf{Query}: \\
\textit{\textcolor{red}{There is} a cat \textcolor{red}{laying down} on a keyboard.}  \\
\textbf{Paraphrased query}: \\
\textit{A cat \textcolor{red}{is stretched out} on a keyboard.}} 
\tabularnewline
\midrule
{\includegraphics[width=\cellwidthbsample cm, height=\cellwidthbsample cm]{samples/000000515577.jpg}} &
\makecell[l]{\textbf{Paraphrasing Score}: 2 \\ 
\textbf{Query}: \\
\textit{Two skiers \textcolor{red}{prepare to make their  way past an embankment}.} \\ 
\textbf{Paraphrased query}: \\
\textit{Two skiers \textcolor{red}{are getting ready to  go around a big hill}.}} 
\tabularnewline
\midrule
{\includegraphics[width=\cellwidthbsample cm, height=\cellwidthbsample cm]{samples/000000003553.jpg}} &
\makecell[l]{\textbf{Paraphrasing Score}: 1 \\
\textbf{Query}: \\
\textit{A person skating on \textcolor{red}{some pieces} of \textcolor{red}{wood}}. \\ 
\textbf{Paraphrased query}:\\ 
\textit{A person skating on \textcolor{red}{a rink made} of \textcolor{red}{ice}}.} 
\tabularnewline
\bottomrule
\end{tabular}
\caption{\textbf{Paraphrased samples by GPT-3.} We show five sample images, their original queries, and the corresponding paraphrases generated by GPT-3. From top to bottom, we show the human-provided semantic similarity scores. A semantic similarity score of 5 indicates that the paraphrase has an identical meaning to the original query.}
    \label{fig:para_samples}
\end{figure}

\begin{figure*}[!t]
    \centering
    \includegraphics[width= .99\linewidth]{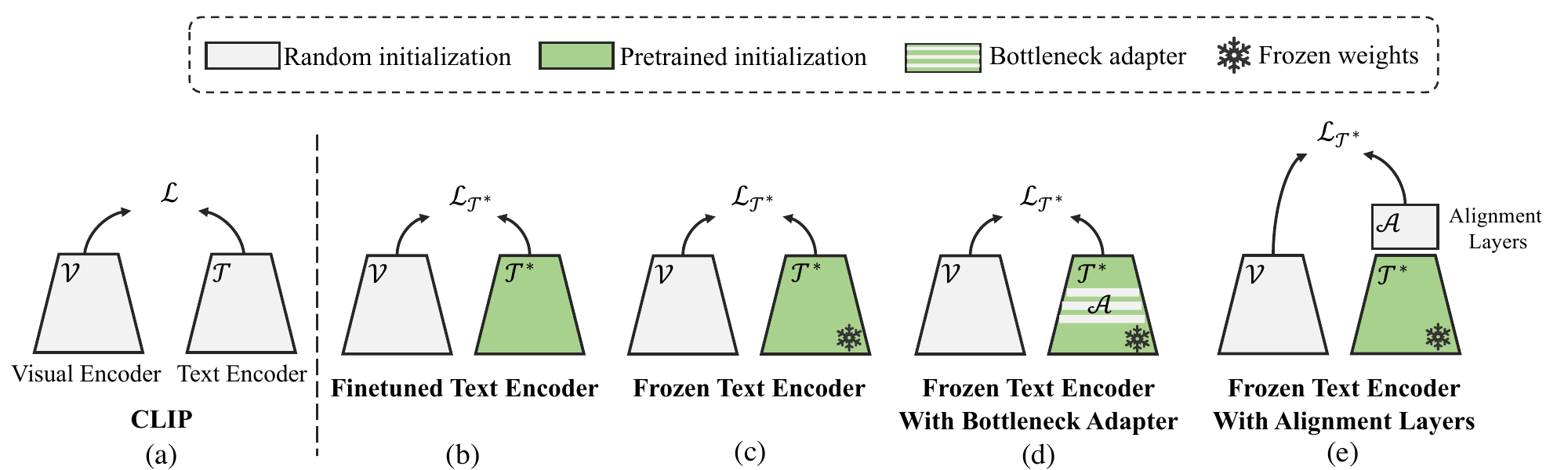}
    \caption{\textbf{Adapting pre-trained language models for paraphrased retrieval.} We explore different strategies to adapt pre-trained language models for paraphrased retrieval. (a) illustrates the CLIP baseline, which trains a dual encoder that optimizes the InfoNCE Loss ($\mathcal{L}$). For CLIP, both the visual encoder ($\mathcal{V}$) and text encoder ($\mathcal{T}$) are trained  with random initialization. (b-e) illustrate the adaptation strategies that we study -- all leveraging a pre-trained language encoder $\mathcal{T}^*$ while training the visual encoder from scratch. This set of adaptation approaches optimize a loss $\mathcal{L}_{\mathcal{T}^*}$. (b) finetunes the text encoder weights, (c) freezes the text encoder weights, (d) inserts bottleneck adapter \cite{bapna-firat-2019-simple,houlsby2019parameter} layers $\mathcal{A}$ to the text encoder, and (e) appends alignment layers, $\mathcal{A}$, atop of the text encoder.}
    \label{fig:adaptation_alternatives}
\end{figure*}

\noindent\textbf{Evaluating the Rank Similarity of Paraphrased Queries.}\label{sec:rank_corr}
In addition to evaluating zero-shot retrieval/classification accuracy of a trained vision-language model, we seek to measure the rank similarity of the top-returned visual search results for a language query and its paraphrase. Typically, it is important for a search engine to show a user the most important search results first. Therefore, an ideal rank similarity metric should satisfy two conditions: (1) it should compare the top-$k$ retrieval list for two queries, and (2) it should weight more heavily the top-most retrieved results. While Spearman's and Kendall's \cite{kendall1948rank} rank correlation coefficients measure the rank similarity for a pair of full retrieval lists, they are unable to handle the aforementioned conditions (retrieval list truncation, top-retrieval weighting). We instead measure (top-$k$ average overlap and top-$k$ Jaccard similarity), which are widely used for computing the similarity of top-$k$ retrieval lists of web search engines \cite{bar2006methods, hannak2013measuring} and recommendation systems \cite{oh2022rank}.  

Let $L_q$ and $L_p$ be rank-ordered retrieval lists for a query $q$ and paraphrase $p$. The \textbf{top-$k$ average overlap} ($\text{AO}@k$) \cite{fagin2003comparing} between the two rank-ordered retrieval lists $L_q, L_p$ is the weighted sum of truncated-list intersections,
\begin{align}
     \text{AO}@k(L_q, L_p) = \frac{1}{k} \sum_{d=1}^k \frac{|L_q^d \cap L_p^d |}{d} \in [0,1],
\end{align}
where $L_q^d=L_q[1:d]$, $L_p^d=L_p[1:d]$ are truncated lists at depth $d$ and $|L_q^d \cap L_p^d |$ denotes the cardinality of the set intersection of the truncated lists. 
$\text{AO}@k=1$ implies that  $L_q^k, L_p^k$ are identical, while $\text{AO}@k=0$ implies that there is no overlap between $L_q^k, L_p^k$. Note that $\text{AO}@k$ is top-weighted as the early retrievals appear in more of the summation terms than the lower retrievals. 

We also measure the \textbf{top-$k$ Jaccard similarity} ($\text{JS}@k$) \cite{jaccard1912distribution}, computed as the ratio of the top-$k$ intersection over union, 
\begin{align}
    \text{JS}@k(L_q, L_p) = \frac{|L_q^k \cap L_p^k|}{|L_q^k \cup L_p^k|} \in [0,1].
\end{align}
$\text{JS}@k$ is minimized when  $L_q^k, L_p^k$ are disjoint and maximized when $L_q^k, L_p^k$ contain the same retrievals (although they may not necessarily be in the same order). Note that the Jaccard similarity can handle retrieval list truncation, but it does not weight top retrievals. 
In general, higher $\text{AO}@k$ and $\text{JS}@k$ indicate that two top-$k$ lists are more similar.

\section{Adapting a Pretrained Language Model for Paraphrased Retrieval} 
\label{sec:methods}

Given a dataset $\mathcal{D}$ of image-text pairs, prior work has trained vision $\mathcal{V}$ and language $\mathcal{T}$ dual encoders from scratch using a contrastive InfoNCE loss $\mathcal{L}$~\cite{oord2018representation},
\begin{equation}
    \min_{\mathcal{V},\mathcal{T}} \mathcal{L}\left(\mathcal{V}, \mathcal{T}; \mathcal{D}\right).
    \label{eqn:clip_loss}
\end{equation}
In this work, we employ the \textbf{CLIP} model~\cite{radford2019language} (illustrated in Figure~\ref{fig:adaptation_alternatives}(a)) as baseline.

As we will show in the experiments, the CLIP model often returns very different retrievals for paraphrased queries. Our hypothesis is that CLIP's learned text encoder does not perform as well on downstream language tasks as state-of-the-art pretrained language models~\cite{devlin-etal-2019-bert,reimers-gurevych-2019-sentence,he2021deberta}. To validate our hypothesis, we aim to train a dual-encoder model starting from a pretrained language model instead of training the language tower from scratch (random initialization). 

In this section, we describe several strategies for learning a dual-encoder model starting from a pretrained language model $\mathcal{T}^\star$. 
Let $\mathcal{L}_{\mathcal{T}^\star}$ be an InfoNCE loss guided by the pretrained language model $\mathcal{T}^\star$. We seek to optimize loss $\mathcal{L}_{\mathcal{T}^\star}$ jointly over the vision encoder $\mathcal{V}$ and a set of auxiliary variables $\mathcal{A}$ in the language tower,
\begin{equation}
    \min_{\mathcal{V},\mathcal{A}} \mathcal{L}_{\mathcal{T}^\star}\left(\mathcal{V}, \mathcal{A}; \mathcal{D}\right).
\end{equation}
Figure~\ref{fig:adaptation_alternatives}(b-e) illustrate the different strategies that we consider in our study, which we describe next. 

\noindent\textbf{Finetuned Text Encoder.} 
This strategy sets the auxiliary variables to be the text encoder $\mathcal{A}\leftarrow\mathcal{T}$ and finetunes starting from the pretrained language model $\mathcal{T}^\star$. As we will show in the experiments, this setting is not optimal for paraphrase retrieval, which is likely due to catastrophic forgetting during training~\cite{kirkpatrick2017overcoming,robins1995catastrophic}.

\noindent\textbf{Frozen Text Encoder.} 
One strategy to avoid catastrophic forgetting in the language tower is to freeze the language encoder to the pretrained model $\mathcal{T}\leftarrow\mathcal{T}^\star$ and only optimize the vision encoder $\mathcal{V}$, \ie, we do not optimize any auxiliary variables ($\mathcal{A}\leftarrow\varnothing$). While this strategy trivially avoids catastrophic forgetting for downstream language tasks, as shown in prior work~\cite{zhai2022lit}, it significantly degrades the downstream zero-shot text-image retrieval and image classification accuracy. Note that this result also holds when projecting the language feature to the joint vision-language embedding space using a linear or multilayer perceptron projection head.

\noindent\textbf{Frozen Text Encoder with Bottleneck Adapter.} 
To overcome the issues with the previous strategies, we seek to increase the adaptation ability of the language encoder while preserving the knowledge from the pretrained language model. To achieve this goal, we freeze the language encoder to the pretrained transformer model $\mathcal{T}\leftarrow\mathcal{T}^\star$ and insert a set of trainable bottleneck adapter modules~\cite{bapna-firat-2019-simple,houlsby2019parameter} $\mathcal{A}$ in each transformer layer. 
In this work, we employed bottleneck adapters with a reduction factor of 2 at all layers of the frozen pretrained text transformer.

\noindent\textbf{Frozen Text Encoder with Alignment Layers.} 
As an alternate adaptation strategy, we also try appending additional transformer layers $\mathcal{A}$ after the frozen text transformer $\mathcal{T}\leftarrow\mathcal{T}^\star$. Since these additional layers are used to align the text embedding with the image embedding, we refer to them as \emph{alignment layers} in this work. In our practice, the alignment layers are set to have the same structure as the frozen text transformer layers with randomly initialized weights.

\section{Experiments}\label{sec:experiments}

The experiment section consists of three parts. Firstly, we ablate various adaptation alternatives and method designs (Section \ref{sec:exp:ablation}). Secondly, we compare our approach to a query expansion baseline in Section \ref{sec:exp:qe}. Lastly, in Section \ref{sec:exp:LS}, we assess the scalability of our final model.  

\noindent \textbf{Evaluation tasks, datasets, and criteria.} 

\noindent\textbf{Paraphrased retrieval.} We evaluate paraphrase ranking similarity on the benchmark proposed in Section \ref{sec:dataset}. The benchmark includes two testing subsets: COCO Paraphrase (COCO-P) and COCO Paraphrase Extended (COCO-PE). We report top-10 Average Overlap (AO@10) and top-10 Jaccard Similarity (JS@10).\\
\noindent\textbf{Classification and retrieval.} We evaluate zero-shot classification on ImageNet 1K (IN-1K) \cite{deng2009imagenet} and report Top-1 accuracy. We also evaluate zero-shot image retrieval on the Flickr30k \cite{plummer2015flickr30k} and COCO Captions (COCO) \cite{lin2014microsoft} datasets. For retrieval, we report the average Recall at 5 (R@5).\\
\noindent\textbf{Text semantic similarity.} We evaluate the zero-shot capabilities of the text encoder on the Semantic Textual Similarity Benchmark (STS-B) \cite{cer-etal-2017-semeval} without finetuning.This dataset is widely used to measure the language understanding ability of language models. We adopt the Spearman \cite{kendall1948rank} and Pearson rank correlation coefficients for measuring the STS performance. 

\medskip
\noindent\textbf{Implementation Details.}
We train our models on the Conceptual Captions 12M (CC12M)~\cite{changpinyo2021conceptual} for ablation study in Sections \ref{sec:exp:ablation} and \ref{sec:exp:qe}. Our visual encoder ($\mathcal{V}$) is randomly initialized and employs a ViT-B/32 architecture~\cite{dosovitskiy2021an} as a backbone. The text encoder ($\mathcal{T}$) adopts the DistilRoBERTa-base \cite{sanh2019distilbert} backbone. This text transformer model has 6 layers, 12 heads, a 768 dimensional embedding space, and $\sim$82M parameters in total (comparable to $\sim$65M parameters for the text encoder of CLIP (ViT-B/32)~\cite{radford2021learning}). We initialize the text encoder leveraging weights provided by the sentence-transformer \cite{reimers-gurevych-2019-sentence}, which were pretrained on the OpenWebText Corpus~\cite{Gokaslan2019OpenWeb} and then finetuned on 1B sentences. We use a linear layer and a 2-layer MLP to project the visual and language embedding to the same multi-modal latent space, respectively. We closely follow the training procedure of CLIP~\cite{radford2021learning} and trained all our models in automatic mixed precision with an AdamW optimizer \cite{loshchilov2018decoupled}, and a cosine annealing learning rate schedule. Our models were trained for 32 epochs including 10000 steps of linear warm-up. Additional hyper-parameters are reported in the supplementary material.

\begin{table*}[t]
\vspace{-1em}
\centering
\subfloat[
\textbf{Paraphrased retrieval}. All adaptation approaches boost ranking similarity (AO@10 and JS@10). Adapting with alignment layers offers the largest gain.
\label{tab:ablation:paraphrase}
]{
\begin{minipage}{0.34\linewidth}{\begin{center}
\tablestyle{4pt}{1.1}
\begin{tabular}{z{62}x{18}x{18}x{18}x{18}}
  & \multicolumn{2}{c}{\scriptsize{COCO-P}} & \multicolumn{2}{c}{\scriptsize{COCO-PE}} \\
  & \scriptsize{AO@10} & \scriptsize{JS@10} & \scriptsize{AO@10} & \scriptsize{JS@10} \\
\shline
\scriptsize{CLIP} & 50.9 & 42.5 & 33.8 & 26.4 \\
\scriptsize{$\mathcal{T}$ \textit{finetuned}} & 51.9 & 44.9 & 34.0 & 26.7 \\
\scriptsize{$\mathcal{T}$ \textit{frozen}} 
& 65.6 & 56.9 & 52.0 & 42.9 \\
\scriptsize{$\mathcal{T}$ \textit{frozen + bottleneck}} 
& 65.5 & 57.0 & 52.7 & 43.1 \\
\scriptsize{$\mathcal{T}$ \textit{frozen + alignment}} 
& \better{68.3} & \better{60.2} & \better{55.2} & \better{46.1} \\

\end{tabular}
\end{center}}\end{minipage}
}
\hspace{2em}
\subfloat[
\textbf{Classification and retrieval.} Adapting with alignment layers yields higher or comparable zero-shot classification (top-1 accuracy) and image retrieval (R@5) performance to the CLIP baseline.
\label{tab:ablation:accuracy}
]{
\centering
\begin{minipage}{0.29\linewidth}{\begin{center}
\tablestyle{2pt}{1.1}
\begin{tabular}{z{62}x{22}x{24}x{24}}
  & \scriptsize{IN-1K} & \scriptsize{Flickr30k} & \scriptsize{COCO} \\
  & \scriptsize{Acc.} & \scriptsize{R@5} & \scriptsize{R@5}\\
\shline
\scriptsize{CLIP} & \better{29.0} & 67.2 & \better{35.8} \\
\scriptsize{$\mathcal{T}$ \textit{finetuned}} & 28.5 & 67.7 & 34.9 \\
\scriptsize{$\mathcal{T}$ \textit{frozen}} & 18.7 & 52.9 & 25.8 \\
\scriptsize{$\mathcal{T}$ \textit{frozen + bottleneck}} & 18.8 & 53.3 & 25.5 \\
\scriptsize{$\mathcal{T}$ \textit{frozen + alignment}} & \better{29.0} & \better{69.7} & \better{35.8} \\

\end{tabular}
\end{center}}\end{minipage}
}
\hspace{2em}
\subfloat[
\textbf{Text semantic similarity.} All adaptation approaches boost the semantic similarity performance. Adapting with alignment layers achieves competitive performance.
\label{tab:ablation:text}
]{
\begin{minipage}{0.24\linewidth}{\begin{center}
\tablestyle{1pt}{1.1}

\begin{tabular}{z{62}x{28}x{28}}
  & \multicolumn{2}{c}{\scriptsize{STS-B}} \\
  & \scriptsize{Pearson} & \scriptsize{Spearman} \\
\shline
\scriptsize{CLIP} & 71.8 & 71.6 \\
\scriptsize{$\mathcal{T}$ \textit{finetuned}} & 72.4 & 72.9 \\
\scriptsize{$\mathcal{T}$ \textit{frozen}} & 82.1 & 83.3 \\
\scriptsize{$\mathcal{T}$ \textit{frozen + bottleneck}} & \better{82.2} & \better{83.4} \\
\scriptsize{$\mathcal{T}$ \textit{frozen + alignment}} & 79.0 & 79.6\\

\end{tabular}
\end{center}}\end{minipage}
}
\\
\centering
\vspace{1em}
\subfloat[
\textbf{Effect of number of alignment layers.} We ablate the number of layers of $\mathcal{T}$ \textit{frozen + alignment}, our best adaptation. Appending layers improves image retrieval but saturates around 6 layers, has a minor impact in paraphrased retrieval (AO@10) and text semantic similarity (Pearson). 
\label{tab:ablation:layers}
]{
\begin{minipage}{0.52\linewidth}{\begin{center}
\tablestyle{2pt}{1.1}
\begin{tabular}{z{108}x{24}x{24}x{24}x{24}x{24}}
  & \multicolumn{5}{c}{\footnotesize{number of alignment layers}} \\
  & \footnotesize{0} & \footnotesize{2} & \footnotesize{4} & \footnotesize{6} & \footnotesize{8} \\
\shline
\footnotesize{retrieval (COCO)} & 25.8 & 33.1 & 35.4 & 35.8 & \better{36.4}\\
\footnotesize{paraphrased retrieval (COCO-P)} & 52.0 & 55.4 & \better{55.5} & 55.2 & 55.3\\
\footnotesize{text semantic similarity} & \better{82.1} & 79.9 & 79.8 & 79.0 & 78.3\\
\end{tabular}
\end{center}}\end{minipage}
}
\hspace{2em}
\subfloat[
\textbf{Effect of initialization of $\mathcal{V}$}. We vary the initialization of the visual encoder between random and ImageNet weights. ImageNet init.\ boosts the retrieval performance but does not improve paraphrased retrieval.
\label{tab:ablation:init}
]{
\centering
\begin{minipage}{0.41\linewidth}{\begin{center}
\tablestyle{4pt}{1.1}
\begin{tabular}{z{106}x{38}x{38}}
  & \multicolumn{2}{c}{\footnotesize{initialization of $\mathcal{V}$}} \\
  & \footnotesize{random} & \footnotesize{ImageNet}\\
\shline
\footnotesize{retrieval (COCO)} & 35.8 & \better{45.9} \\
\footnotesize{paraphrased retrieval (COCO-P)} & \better{55.2} & 55.1 \\
\footnotesize{text semantic similarity} & \better{79.0} & 78.9 \\
\end{tabular}
\end{center}}\end{minipage}
}
\hspace{2em}

\caption{\textbf{Adaptation on CC12M ablation studies}. We trained the CLIP baseline and our adaptation variants on the CC12M dataset. We conduct all experiments in a zero-shot fashion. See the text for a discussion of the results.}
\label{tab:ablations}
\end{table*}

\subsection{Ablation Study: Adaptation on CC12M}
\label{sec:exp:ablation}

\noindent\textbf{Adaptation strategies.}
We investigate different adaptation strategies for  training a dual-encoder model starting from a pretrained text encoder. We evaluate the CLIP baseline~\cite{radford2021learning} (Figure \ref{fig:adaptation_alternatives}(a)) and the four different adaptation strategies introduced in Section \ref{sec:methods}: 

    \noindent\textbf{$\mathcal{T}$ \textit{finetuned}} (Figure \ref{fig:adaptation_alternatives}(b)).  Finetunes a pre-trained text encoder without additional trainable layers.\\
    \noindent\textbf{$\mathcal{T}$ \textit{frozen}} (Figure \ref{fig:adaptation_alternatives}(c)). Freezes a pre-trained text encoder without additional trainable layers.\\
    \noindent\textbf{$\mathcal{T}$ \textit{frozen + bottleneck}} (Figure \ref{fig:adaptation_alternatives}(d)). Inserts  bottleneck adapters \cite{houlsby2019parameter,bapna-firat-2019-simple} to a pre-trained (and frozen) text encoder.\\
    \noindent\textbf{$\mathcal{T}$ \textit{frozen + alignment}} (Figure \ref{fig:adaptation_alternatives}(e)). Appends $m=6$ alignment layers to a pre-trained 
 (and frozen) text encoder.

We present the most important comparisons in Table~\ref{tab:ablations} and include the full ablation results in the supplementary material (Tables \textcolor{red}{B.2} \& \textcolor{red}{B.3}). 

\noindent\textbf{(a) Effect on paraphrased retrieval (Table \ref{tab:ablation:paraphrase}).}\\
We observe that approaches with a frozen pretrained text encoder always achieve significantly higher paraphrase ranking similarity. They outperform significantly the CLIP baseline. On the contrary, with the same pre-trained initialization, the paraphrase ranking similarity of $\mathcal{T}$ \textit{finetuned} drops to the level of CLIP. We attribute that result to catastrophic forgetting of the rich text pre-training, and hypothesize that such forgetting emerges from fine-tuning on millions of noisy image-text pairs.

\noindent\textbf{(b) Effect on classification and retrieval (Table \ref{tab:ablation:accuracy}).}\\
First, we observe that the frozen text encoder has significantly worse retrieval/classification accuracy than CLIP, which is consistent with the observation reported in LiT~\cite{zhai2022lit}. The introduction of the bottleneck adapters failed to make up the accuracy gap. Meanwhile, the approaches of finetuning and introducing alignment layers are shown to achieve comparable, and sometimes even better, zero-shot accuracy than the baseline CLIP method. Note that our finetuning result is consistent with the result reported in LiT~\cite{zhai2022lit}.

\begin{table}[]
\small
    \centering
    \renewcommand{\arraystretch}{1.2}
    \begin{tabular}{rccc}
        &  & COCO & COCO-P  \\
        \midrule
        CLIP (w/o QE) & & 35.8 & 50.9  \\
        CLIP + QE w/ WordNet &  & 34.8 & 52.4  \\
        CLIP + QE w/ Parrot &  & \better{36.7} & 53.9  \\ 
       \hline
        Ours (w/o QE)& & 35.8 & \better{68.3} \\
    \end{tabular}
    \caption{\textbf{Comparison to Query Expansion (QE).} We evaluate two query expansion techniques on image retrieval (R@5) and paraphrased retrieval (AO@10). We compare to our method ($\mathcal{T}$ \textit{frozen + alignment}) and observe that our solution achieves significantly higher paraphrased retrieval (COCO-P). All models are trained on CC12M.}
    \label{tab:QE}
\end{table}

\noindent\textbf{(c) Effect on text semantic similarity (Table \ref{tab:ablation:text}).}\\ 
We employ the text encoder of each adaptation strategy to the semantic textual similarity task on the STS-B dataset. $\mathcal{T}$ \textit{frozen} achieves high scores, which is expected given that the language model is trained on a text corpora. Surprisingly, the $\mathcal{T}$ \textit{frozen + bottleneck} language encoder slightly outperforms the $\mathcal{T}$ \textit{frozen} language model. While $\mathcal{T}$ \textit{frozen + alignment} has slightly lower scores, it is significantly better than CLIP and $\mathcal{T}$ \textit{finetuned}.

\begin{table*}[]
\setlength{\tabcolsep}{5pt}
\begin{tabular}{*l ^c | ^c ^c ^c ^c | ^c ^c ^c ^c | ^c ^c}
& &
\multicolumn{4}{c|}{\scriptsize \textbf{paraphrased retrieval}} &
\multicolumn{4}{c|}{\scriptsize \textbf{classification and retrieval}} &
\multicolumn{2}{c}{\scriptsize \textbf{text semantic similarity}}
\\
& &
\multicolumn{4}{c|}{\scriptsize paraphrased text-image retrieval} &
\multicolumn{2}{c|}{\scriptsize classification} & 
\multicolumn{2}{c|}{\scriptsize image retrieval} & 
\multicolumn{2}{c}{\scriptsize text semantic similarity} \\
 &  & 
 \multicolumn{2}{c}{\scriptsize COCO-P} & \multicolumn{2}{c|}{\scriptsize COCO-PE} &
\scriptsize IN-1K & \multicolumn{1}{c|}{\scriptsize VTAB+} & 
\scriptsize Flickr30k & \multicolumn{1}{c|}{\scriptsize COCO} &
\multicolumn{2}{c}{\scriptsize STS-B}\\
model & data &
\scriptsize AO@10 & \scriptsize JS@10 &
\scriptsize AO@10 & \scriptsize JS@10 &
\multicolumn{2}{c|}{\scriptsize Acc.} &
\multicolumn{2}{c|}{\scriptsize R@5} &
\scriptsize{Pearson} & \scriptsize{Spearman} \\
\shline
CLIP~\cite{radford2021learning} & WIT-400M &
67.0 & 58.1 &
52.8 & 42.6 &
\better{63.3} & 45.5 & 
83.4 & 56.0 &
65.6 & 69.5
\\
OpenCLIP \cite{cherti2022reproducible} & LAION-400M &
68.1 & 59.5 &
51.5 & 42.5 &
62.9 & 45.8 &
\better{85.5} & \better{60.9} &
75.2 & 75.7
\\
\hline
Ours & LAION-400M  &
\better{73.7} & \better{66.1} &
\better{60.2} & \better{50.8} &
60.5 & \better{46.6} &
84.0 & 60.0 &
\better{80.3} & \better{81.0}
\\
\end{tabular}

\caption{\textbf{Large Scale Adaptation on LAION-400M.} We compare our adaptation method ($\mathcal{T}$ \textit{frozen + alignment}) with baselines, all trained at large scale, and evaluated on multiple zero-shot tasks: paraphrased retrieval, classification and retrieval, and text semantic similarity. All models use the ViT-B/32 architecture as the visual backbone and are trained for 32 epochs on the specified dataset. For VTAB+, we report the average top-1 accuracy over 35 tasks following the setup in \cite{schuhmann2022laionb}.
}
\label{tab:scaling}
\end{table*}

\noindent\textbf{(d) Effect of the number of alignment layers (Table \ref{tab:ablation:layers}).}\\ 
Through ablations (a-c), we found that the strategy of appending alignment layers to a frozen pre-trained text encoder ($\mathcal{T}$ \textit{frozen + alignment}) achieves the best balance between retrieval accuracy, paraphrased ranking similarity, and text semantic similarity performance. Therefore, in this analysis we study the effect of the number of alignment layers of the aforementioned model. We observe that increasing the number of layers is beneficial for all tasks, with performance saturating after 6 layers.

\noindent\textbf{(e) Effect of visual encoder initialization (Table \ref{tab:ablation:init}).}\\
We observe that initializing the visual encoder ($\mathcal{V}$) with ImageNet greatly improves the retrieval performance (R@5 on COCO) compared to random initialization. The performance on paraphrased retrieval and text semantic similarity remains relatively unchanged after varying the visual encoder initialization. 

\subsection{Comparison to Query Expansion}
\label{sec:exp:qe}
Query Expansion (QE) is a popular technique to reduce the effect of synonyms on text queries \cite{carpineto2012survey,azad2019query}. Accordingly, we evaluate QE on our paraphrased image retrieval benchmark. We consider two different QE approaches: QE via WordNet \cite{miller1995wordnet} and QE via automatic paraphrasing. The former expands a sentence query by replacing one of its non-stop words with a synonym provided by WordNet. The latter expands sentence queries by generating automatic paraphrases leveraging the T5-based \cite{2020t5} Parrot paraphrasing model \cite{prithivida2021parrot}. Given a query sentence $q^{\text{txt}}$ and $K$ expanded queries $\{q^{\text{txt}}_{i}\}_{i=1}^{K}$, our goal is to obtain a QE-expanded text feature $\textbf{z}^{new}$, and then use it to compute the cosine similarity against the visual features in the retrieval set. This new text feature is formed by averaging the text feature of the original query and the $K$ expanded queries, 
\begin{align}
\textbf{z}^{new} = \frac{1}{K+1} \left(\mathcal{T}(q^{\text{txt}}) + \sum_{i=1}^{K} \mathcal{T}(q^{\text{txt}}_{i})\right).
\end{align}

The results of combining query expansion and the CLIP \textit{baseline} model in Section \ref{sec:exp:ablation} are listed in Table \ref{tab:QE}. Both QE techniques slightly improve ranking similarity in the paraphrased retrieval task. However, their effectiveness is limited compared to our adaptation approach. In addition, we observe a drop in image retrieval accuracy for QE w/ WordNet. We suspect this drop is due to the use of incorrect synonyms resulting from imperfect word sense disambiguation. In light of these results, we argue that QE (alone) is not sufficient for addressing the limitations of CLIP's dual encoder for the paraphrase retrieval task.

\subsection{Large-scale Adaptation on LAION-400M}
\label{sec:exp:LS}

\begin{figure*}[!t]
\begin{center}
\setlength{\tabcolsep}{1.5pt} 
\renewcommand{\arraystretch}{1.0}
\begin{tabular}{ccccccccccc}
\toprule
\multicolumn{11}{c}{{Query: \textit{“A road sign that indicates route \textcolor{red}{sixty six}.”}}} 
\tabularnewline
\highlight{\includegraphics[width=\cellwidth cm, height=\cellwidth cm]{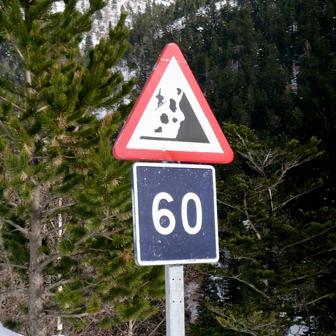}} & 
{\highlight{\includegraphics[width=\cellwidth cm, height=\cellwidth cm]{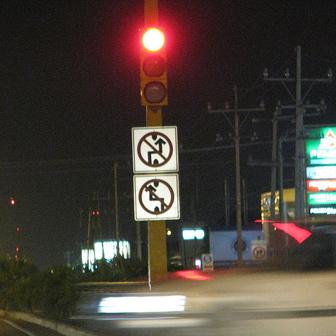}}} &
{\includegraphics[width=\cellwidth cm, height=\cellwidth cm]{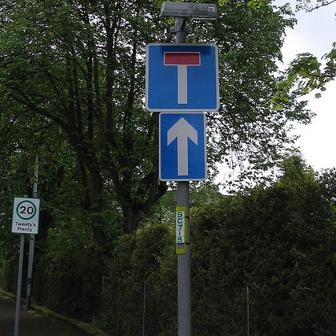}}& 
{\includegraphics[width=\cellwidth cm, height=\cellwidth cm]{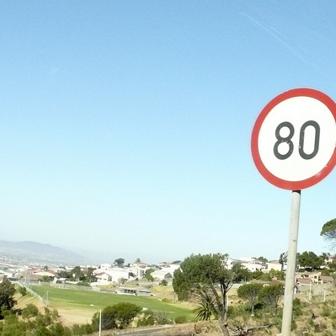}}&
{\includegraphics[width=\cellwidth cm, height=\cellwidth cm]{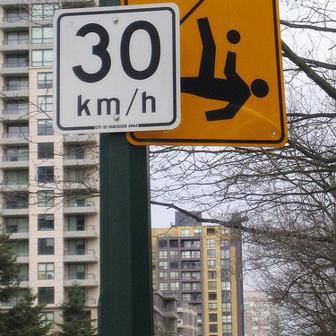}}&
\hspace{1cm} &
{\highlight{\includegraphics[width=\cellwidth cm, height=\cellwidth cm]{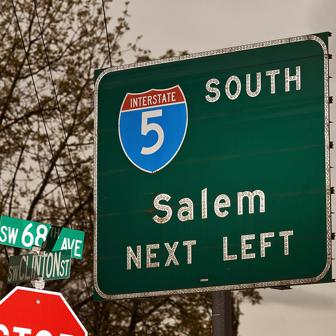}}}&
{\highlight{\includegraphics[width=\cellwidth cm, height=\cellwidth cm]{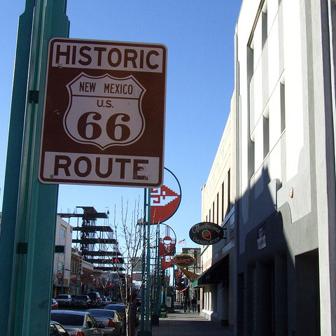}}}&
{\highlight{\includegraphics[width=\cellwidth cm, height=\cellwidth cm]{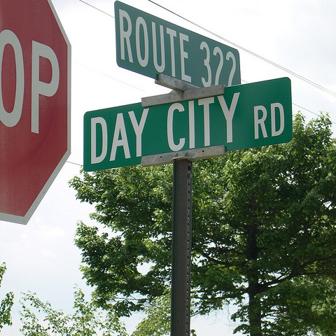}}}&
{\highlight{\includegraphics[width=\cellwidth cm, height=\cellwidth cm]{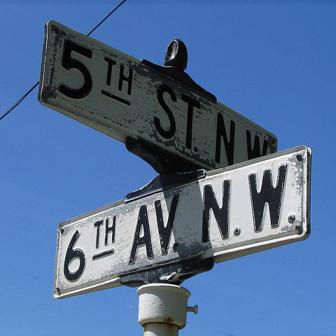}}}&
{\includegraphics[width=\cellwidth cm, height=\cellwidth cm]{samples/resized/3024/000000101022.jpg}}
\tabularnewline
\multicolumn{11}{c}{{Paraphrased query: \textit{"A road sign that indicates Route \textcolor{red}{66}"}}} 
\tabularnewline
{\includegraphics[width=\cellwidth cm, height=\cellwidth cm]{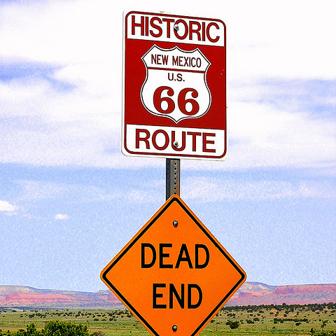}} & 
{\includegraphics[width=\cellwidth cm, height=\cellwidth cm]{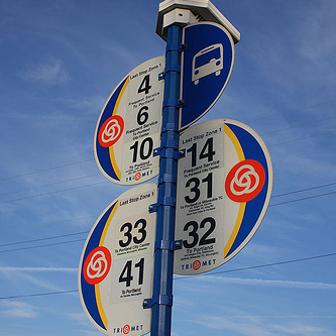}} &
{\highlight{\includegraphics[width=\cellwidth cm, height=\cellwidth cm]{samples/resized/3024/000000108536.jpg}}}& 
{\highlight{\includegraphics[width=\cellwidth cm, height=\cellwidth cm]{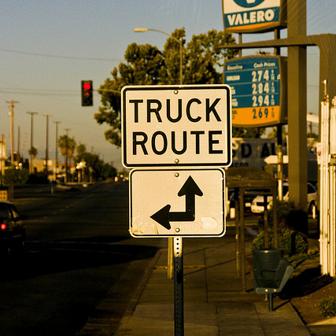}}}&
{\highlight{\includegraphics[width=\cellwidth cm, height=\cellwidth cm]{samples/resized/3024/000000101022.jpg}}}& 
~ &
{\highlight{\includegraphics[width=\cellwidth cm, height=\cellwidth cm]{samples/resized/3024/000000059250.jpg}}}&
{\highlight{\includegraphics[width=\cellwidth cm, height=\cellwidth cm]{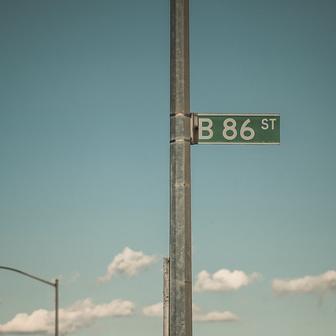}}}&
{\highlight{\includegraphics[width=\cellwidth cm, height=\cellwidth cm]{samples/resized/3024/000000189735.jpg}}}&
{\highlight{\includegraphics[width=\cellwidth cm, height=\cellwidth cm]{samples/resized/3024/000000029287.jpg}}}&
{\includegraphics[width=\cellwidth cm, height=\cellwidth cm]{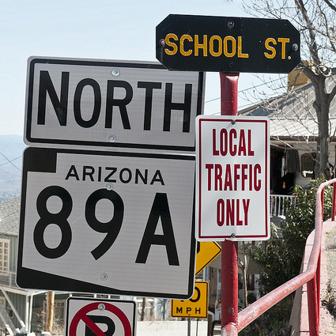}}
\tabularnewline
\midrule
\multicolumn{11}{c}{{Query: \textit{“\textcolor{red}{An} open \textcolor{red}{refrigerator filled with lots} of food.”}}} 
\tabularnewline
{\highlight{\includegraphics[width=\cellwidth cm, height=\cellwidth cm]{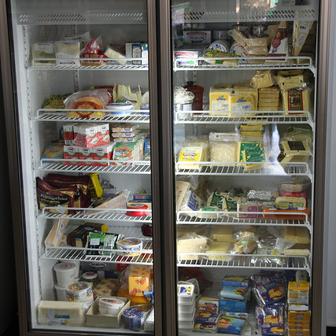}}}& 
{\includegraphics[width=\cellwidth cm, height=\cellwidth cm]{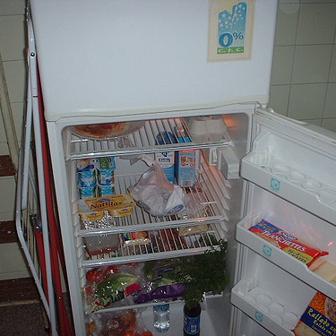}}&
{\highlight{\includegraphics[width=\cellwidth cm, height=\cellwidth cm]{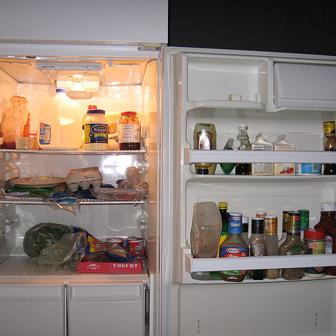}}}& 
{\includegraphics[width=\cellwidth cm, height=\cellwidth cm]{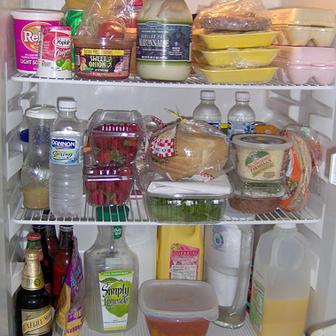}}&
{\includegraphics[width=\cellwidth cm, height=\cellwidth cm]{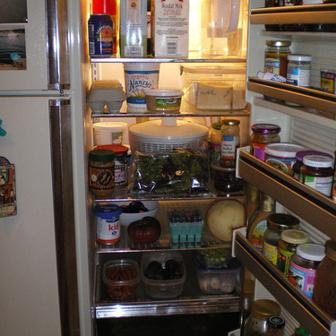}}&
~&
{\highlight{\includegraphics[width=\cellwidth cm, height=\cellwidth cm]{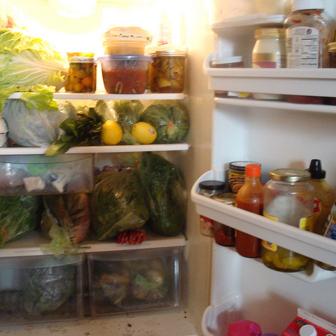}}}&
{\highlight{\includegraphics[width=\cellwidth cm, height=\cellwidth cm]{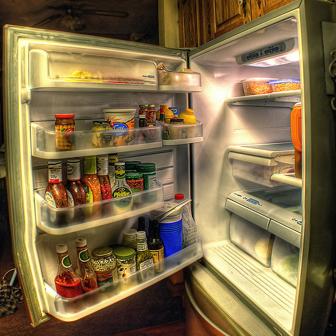}}}&
{\highlight{\includegraphics[width=\cellwidth cm, height=\cellwidth cm]{samples/resized/refrigerator/000000555156.jpg}}}& 
{\highlight{\includegraphics[width=\cellwidth cm, height=\cellwidth cm]{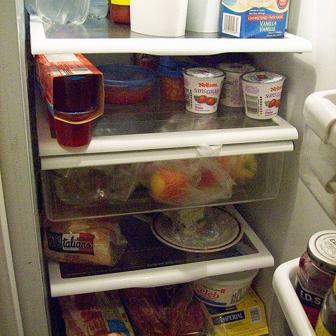}}}&
{\includegraphics[width=\cellwidth cm, height=\cellwidth cm]{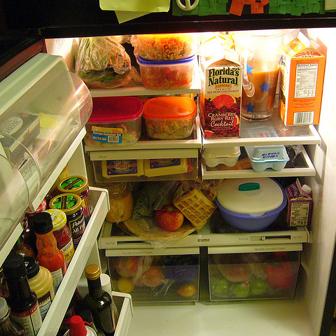}}
\tabularnewline
\multicolumn{11}{c}{{Paraphrased query: \textit{"\textcolor{red}{A fridge that} is open \textcolor{red}{and has a lot} of food \textcolor{red}{inside}"}}} 
\tabularnewline
{\includegraphics[width=\cellwidth cm, height=\cellwidth cm]{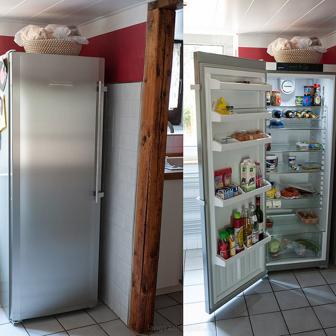}}&
{\highlight{\includegraphics[width=\cellwidth cm, height=\cellwidth cm]{samples/resized/refrigerator/000000234328.jpg}}}&
{\includegraphics[width=\cellwidth cm, height=\cellwidth cm]{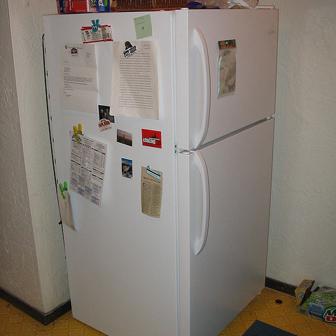}}& 
{\includegraphics[width=\cellwidth cm, height=\cellwidth cm]{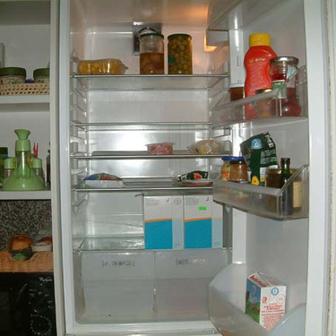}}&
{\includegraphics[width=\cellwidth cm, height=\cellwidth cm]{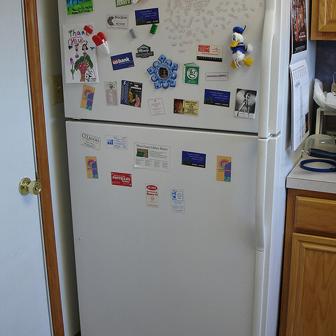}}&
~&
{\highlight{\includegraphics[width=\cellwidth cm, height=\cellwidth cm]{samples/resized/refrigerator/000000064704.jpg}}}&
{\highlight{\includegraphics[width=\cellwidth cm, height=\cellwidth cm]{samples/resized/refrigerator/000000555156.jpg}}}&
{\highlight{\includegraphics[width=\cellwidth cm, height=\cellwidth cm]{samples/resized/refrigerator/000000541856.jpg}}}& 
{\highlight{\includegraphics[width=\cellwidth cm, height=\cellwidth cm]{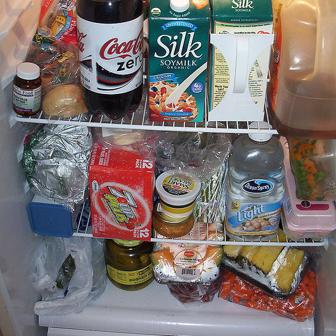}}}&
{\includegraphics[width=\cellwidth cm, height=\cellwidth cm]{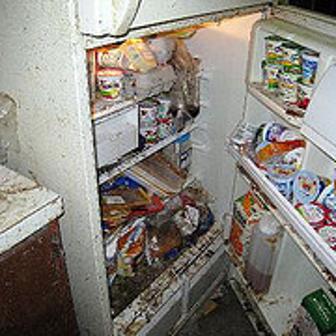}}
\tabularnewline
\midrule
\multicolumn{11}{c}{{Query: \textit{“A cat \textcolor{red}{sitting on the back of} a sofa \textcolor{red}{looking} out a window.”}}} 
\tabularnewline
{\includegraphics[width=\cellwidth cm, height=\cellwidth cm]{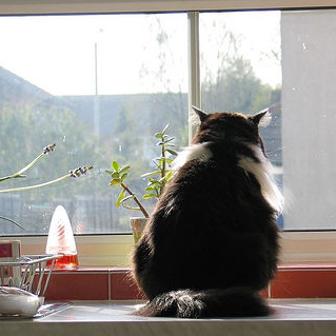}}&
{\includegraphics[width=\cellwidth cm, height=\cellwidth cm]{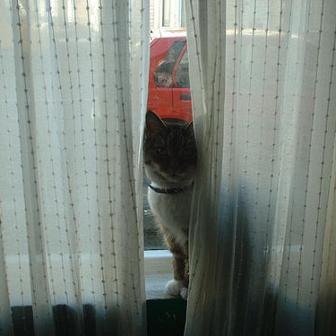}}&
{\highlight{\includegraphics[width=\cellwidth cm, height=\cellwidth cm]{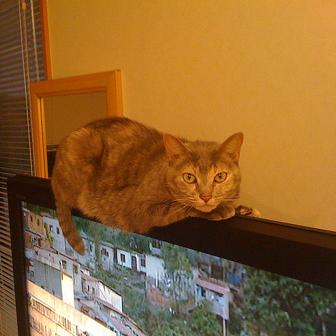}}}& 
{\highlight{\includegraphics[width=\cellwidth cm, height=\cellwidth cm]{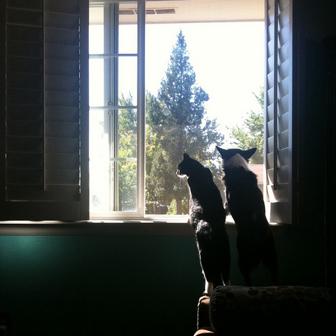}}}&
{\includegraphics[width=\cellwidth cm, height=\cellwidth cm]{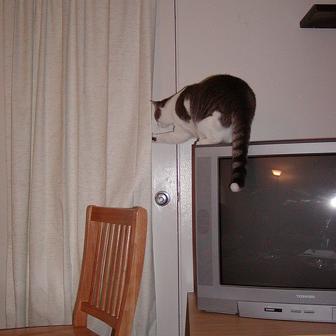}}&
~&
{\highlight{\includegraphics[width=\cellwidth cm, height=\cellwidth cm]{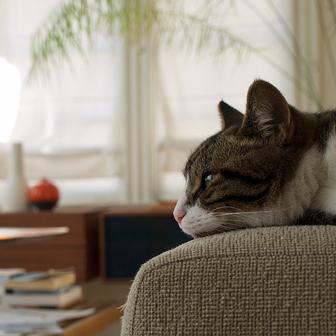}}}&
{\highlight{\includegraphics[width=\cellwidth cm, height=\cellwidth cm]{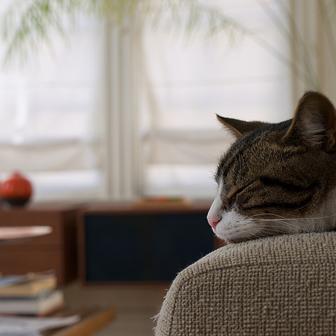}}}&
{\highlight{\includegraphics[width=\cellwidth cm, height=\cellwidth cm]{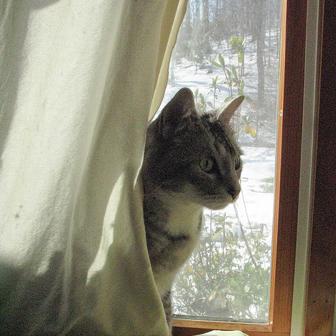}}}& 
{\highlight{\includegraphics[width=\cellwidth cm, height=\cellwidth cm]{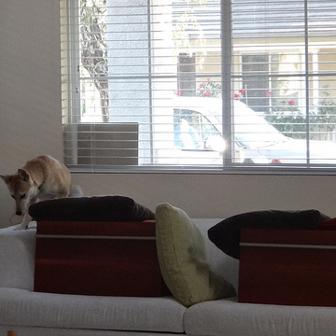}}}&
{\includegraphics[width=\cellwidth cm, height=\cellwidth cm]{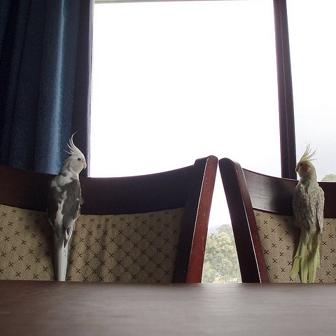}}
\tabularnewline
\multicolumn{11}{c}{{Paraphrased query: \textit{A cat \textcolor{red}{perched atop} a sofa {gazing} out a window.}}} 
\tabularnewline
{\highlight{\includegraphics[width=\cellwidth cm, height=\cellwidth cm]{samples/resized/3819/000000563869.jpg}}}&
{\highlight{\includegraphics[width=\cellwidth cm, height=\cellwidth cm]{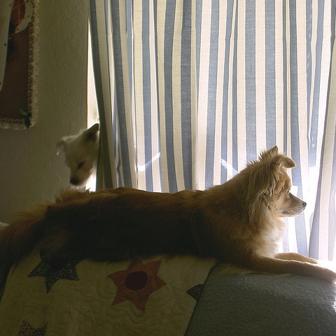}}}&
{\highlight{\includegraphics[width=\cellwidth cm, height=\cellwidth cm]{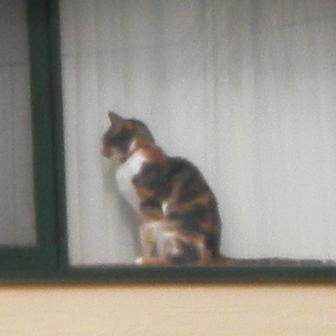}}}& 
{\includegraphics[width=\cellwidth cm, height=\cellwidth cm]{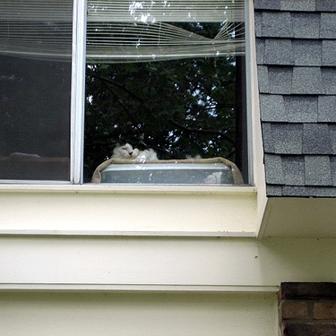}}&
{\includegraphics[width=\cellwidth cm, height=\cellwidth cm]{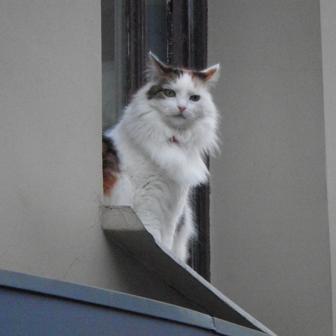}}&
~&
{\highlight{\includegraphics[width=\cellwidth cm, height=\cellwidth cm]{samples/resized/3819/000000342078.jpg}}}&
{\highlight{\includegraphics[width=\cellwidth cm, height=\cellwidth cm]{samples/resized/3819/000000070353.jpg}}}&
{\highlight{\includegraphics[width=\cellwidth cm, height=\cellwidth cm]{samples/resized/3819/000000226220.jpg}}}& 
{\highlight{\includegraphics[width=\cellwidth cm, height=\cellwidth cm]{samples/resized/3819/000000237419.jpg}}}&
{\highlight{\includegraphics[width=\cellwidth cm, height=\cellwidth cm]{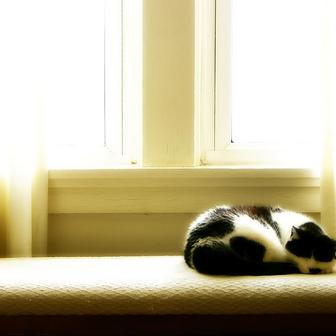}}}
\tabularnewline
\midrule
\multicolumn{11}{c}{{Query: \textit{“A toy horse \textcolor{red}{sits atop} a red wooden chair.”}}} 
\tabularnewline
{\highlight{\includegraphics[width=\cellwidth cm, height=\cellwidth cm]{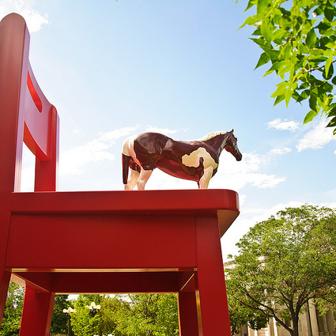}}}&
{\highlight{\includegraphics[width=\cellwidth cm, height=\cellwidth cm]{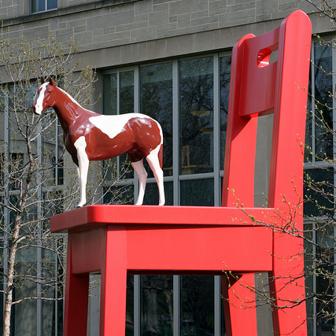}}}&
{\highlight{\includegraphics[width=\cellwidth cm, height=\cellwidth cm]{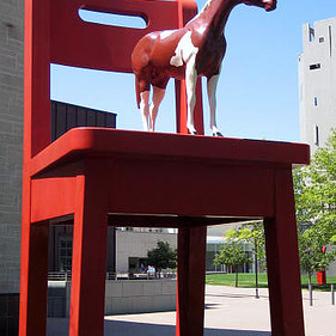}}}& 
{\highlight{\includegraphics[width=\cellwidth cm, height=\cellwidth cm]{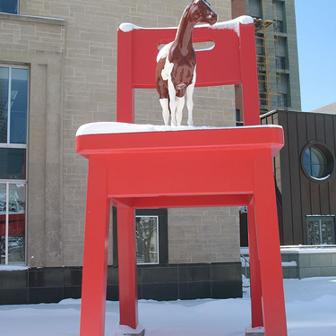}}}&
{\highlight{\includegraphics[width=\cellwidth cm, height=\cellwidth cm]{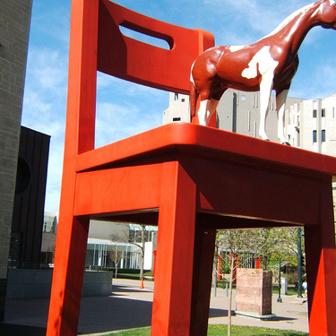}}}&
~&
{\highlight{\includegraphics[width=\cellwidth cm, height=\cellwidth cm]{samples/resized/1494/000000004979.jpg}}}&
{\highlight{\includegraphics[width=\cellwidth cm, height=\cellwidth cm]{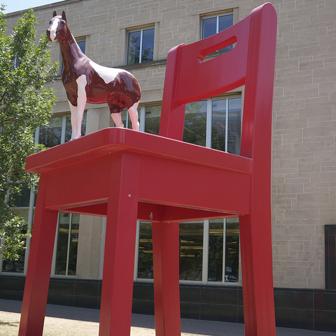}}}&
{\includegraphics[width=\cellwidth cm, height=\cellwidth cm]{samples/resized/1494/000000269932.jpg}}& 
{\highlight{\includegraphics[width=\cellwidth cm, height=\cellwidth cm]{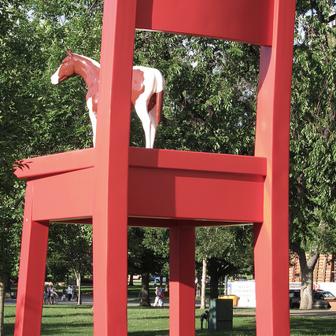}}}&
{\includegraphics[width=\cellwidth cm, height=\cellwidth cm]{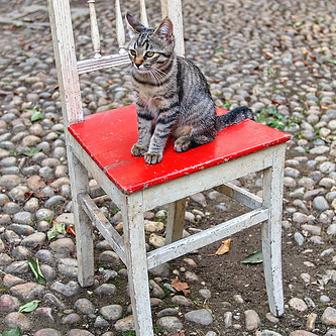}}
\tabularnewline
\multicolumn{11}{c}{{Paraphrased query: \textit{“A toy horse \textcolor{red}{is perched on top of} a red wooden chair.”}}} 
\tabularnewline
{\highlight{\includegraphics[width=\cellwidth cm, height=\cellwidth cm]{samples/resized/1494/000000127576.jpg}}}&
{\highlight{\includegraphics[width=\cellwidth cm, height=\cellwidth cm]{samples/resized/1494/000000468017.jpg}}}&
{\highlight{\includegraphics[width=\cellwidth cm, height=\cellwidth cm]{samples/resized/1494/000000004979.jpg}}}& 
{\highlight{\includegraphics[width=\cellwidth cm, height=\cellwidth cm]{samples/resized/1494/000000413414.jpg}}}&
{\highlight{\includegraphics[width=\cellwidth cm, height=\cellwidth cm]{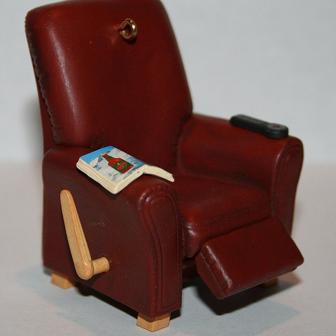}}}&
~&
{\highlight{\includegraphics[width=\cellwidth cm, height=\cellwidth cm]{samples/resized/1494/000000276209.jpg}}}&
{\highlight{\includegraphics[width=\cellwidth cm, height=\cellwidth cm]{samples/resized/1494/000000004979.jpg}}}&
{\includegraphics[width=\cellwidth cm, height=\cellwidth cm]{samples/resized/1494/000000468017.jpg}}& 
{\highlight{\includegraphics[width=\cellwidth cm, height=\cellwidth cm]{samples/resized/1494/000000407220.jpg}}}&
{\includegraphics[width=\cellwidth cm, height=\cellwidth cm]{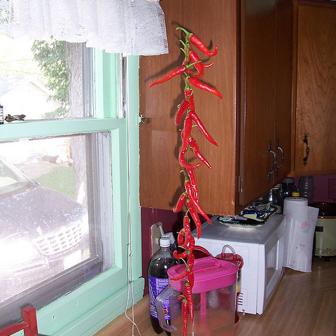}}
\tabularnewline
\bottomrule
\multicolumn{5}{c}{{CLIP's top retrievals}} & ~ &  \multicolumn{5}{c}{{Our top retrievals}}
\end{tabular}
\end{center}
\vspace{-0.2cm}
\caption{\textbf{Qualitative results} for CLIP~\cite{radford2021learning} (left) and our approach (right). We outline images in green that appear in the top-10 retrievals for both the query and paraphrased query.  The full top-10 retrievals can be found in the supplementary material (Figures \textcolor{red}{B.1} \& \textcolor{red}{B.2}). Please see the text for a discussion of the results.}
    \label{fig:qualitative}
\end{figure*}

In this section, we study the performance of our best model, Frozen Text Encoder with Alignment Layers ($\mathcal{T}$ \textit{frozen + alignment}), trained on a large-scale dataset. We train our model using the LAION-400M dataset \cite{schuhmann2021laion}, which contains about 413M image-text pairs. Our model's visual backbone is randomly initialized. We compare our method with CLIP (ViT-B/32)~\cite{radford2021learning} model and OpenCLIP (ViT-B/32)~\cite{ilharco_gabriel_2021_5143773}. Note that CLIP employs a private dataset, while OpenCLIP leverages the same training dataset as our adaptation approach. The training total batch size is set as 32760. Our model is trained for 32 epochs using 40 NVIDIA A100 80GB GPUs in one week.  For CLIP and OpenCLIP, we use their publicly available model weights.

Table \ref{tab:scaling} summarizes the results on three different tasks. In this setting, we added the VTAB+ \cite{zhai2019large,schuhmann2022laionb} benchmark as an additional comparison for classification. This benchmark includes 35 different datasets, and has become widely adopted to evaluate the zero-shot capabilities of large scale pre-trained models. Consistently with the results in Section \ref{sec:exp:ablation}, our model improves upon the CLIP and OpenCLIP baselines on the paraphrased retrieval task. As an example, our model achieves 60.2\% $\text{AO}@10$ on COCO-PE, which represents a +7.4\% gain with respect to the most competitive baseline (CLIP). For classification and retrieval, our model achieves competitive/comparable performance for most of the datasets, and improves +0.8\% on VTAB+, arguably one of the most challenging benchmarks. Finally, we also observe that the text encoder of our adapted model is much more appealing for text semantic similarity. Our model improves over the most competitive baseline (OpenCLIP) by 5.1\% on this task.

\begin{table}[t]
\centering
\small
\begin{tabular}{*l | c | c }
& COCO  & COCO-TP
\\
\midrule
\multirow{1}{*}{CLIP \cite{radford2021learning}}  & 361.8  & 308.1 ( -14.8\% )
\\
\multirow{1}{*}{OpenCLIP \cite{cherti2022reproducible}}  & 381.6  & 329.3 ( -13.7\% )
\\
\midrule
\multirow{1}{*}{Ours}  & 371.6  & \better{335.3 ( -9.8\% )}  
\\
\end{tabular}

\caption{Evaluations of Robustness to text perturbations on COCO-TP \cite{qiu2022multimodal}. RSUM (\ie, $\sum_{k\in\{1,5,10\}} \text{IR R@$k$} + \text{TR R@$k$}$ is employed as evaluation metrics.} 
\label{tab:COCOTP}
\end{table}

In Figure~\ref{fig:qualitative}, we show qualitative results of our approach and compare against CLIP. Notice how, in general, the top retrievals for our approach has more images that are in common between the query and paraphrased query (outlined in green) than the CLIP baseline. Moreover, our approach handles phrase synonyms (first, third examples) and full sentence paraphrases (second example). The bottom result shows a failure case where our approach underperforms CLIP, likely due to forgetting during training.

In addition to the evaluation discussed above, we compared our model with CLIP and OpenCLIP on COCO-TP~\cite{qiu2022multimodal}, a benchmark of the robustness of vision-language models to different types of text perturbations. Higher robustness of our model to text perturbations is observed in Table \ref{tab:COCOTP} and can be considered as a beneficial side effect of our adaption strategy. 

\section{Conclusion}
\label{sec:conclusion}
We have demonstrated an effective strategy for training a dual-encoder vision-language model by adapting from a pretrained language model. Our strategy yields significantly improved paraphrased retrieval on our newly collected paraphrased text-image dataset. Our dual encoder's text tower also improves on the downstream text semantic similarity task. Moreover, our strategy maintains downstream zero-shot classification and retrieval accuracy. Our work opens up the possibility of an improved natural language search interface and a tighter coupling between strong language models and learned visual representations.



\clearpage

{
    \small
    \bibliographystyle{ieeenat_fullname}
    \bibliography{main}
}

\clearpage
\appendix
\onecolumn

\counterwithin{figure}{section}
\counterwithin{table}{section}


\begin{center}
{\bf \Large	 
Adapting Dual-encoder Vision-language Models for Paraphrased Retrieval
\bigskip Supplementary Material
}
\end{center}

\section{Paraphrased Text-Image Retrieval Dataset}

\subsection{Samples.}
We present 10 samples from our paraphrased text-image retrieval dataset in Figure \ref{supp:fig:para_samples}. 

\begin{figure*}[!h]
\centering
\small
\begin{tabular}{|m{1.2cm}|l|}
\toprule
\centering{Image} & \centering{Caption}  
\tabularnewline
\midrule
{\includegraphics[width=\cellwidthsupp cm, height=\cellwidthsupp cm]{samples/000000049810.jpg}} &
\makecell[l]{Original Query: \textit{A \textcolor{red}{little} cat \textcolor{red}{looking} at \textcolor{red}{itself} in a mirror.} \\
Paraphrased Query: \textit{A \textcolor{red}{small} cat \textcolor{red}{gazing} at \textcolor{red}{its reflection} in a mirror.}}
\tabularnewline
\midrule


{\includegraphics[width=\cellwidthsupp cm, height=\cellwidthsupp cm]{samples/000000288584.jpg}} &
\makecell[l]{Original Query: \textit{A person \textcolor{red}{dressed as} a giraffe carrying a \textcolor{red}{bullhorn}.} \\ Paraphrased Query: \textit{A person \textcolor{red}{wearing} a giraffe \textcolor{red}{costume} and carrying a \textcolor{red}{megaphone}.}}
\tabularnewline
\midrule
{\includegraphics[width=\cellwidthsupp cm, height=\cellwidthsupp cm]{samples/000000287959.jpg}} &
\makecell[l]{Original Query: \textit{a sandwich that \textcolor{red}{has} something green and red \textcolor{red}{in it}} \\ Paraphrased Query: \textit{A sandwich that \textcolor{red}{contains} something green and red.}}
\tabularnewline
\midrule
{\includegraphics[width=\cellwidthsupp cm, height=\cellwidthsupp cm]{samples/000000527695.jpg}} &
\makecell[l]{Original Query: \textit{A bowl of some broccoli \textcolor{red}{that is} chopped.} \\ Paraphrased Query: \textit{A bowl of some chopped broccoli.}}
\tabularnewline
\midrule
{\includegraphics[width=\cellwidthsupp cm, height=\cellwidthsupp cm]{samples/000000303305.jpg}} &
\makecell[l]{Original Query: \textit{A \textcolor{red}{large} red and white bus on \textcolor{red}{a city street}.} \\ Paraphrased Query: \textit{A \textcolor{red}{big} red and white bus on \textcolor{red}{an urban road}.}}
\tabularnewline
\midrule
{\includegraphics[width=\cellwidthsupp cm, height=\cellwidthsupp cm]{samples/000000518770.jpg}} &
\makecell[l]{Original Query: \textit{\textcolor{red}{Some kids} standing up playing \textcolor{red}{wii with each other}} \\ Paraphrased Query: \textit{\textcolor{red}{A group of children} are playing \textcolor{red}{the Wii console together while} standing up.}}
\tabularnewline
\midrule
{\includegraphics[width=\cellwidthsupp cm, height=\cellwidthsupp cm]{samples/000000549220.jpg}} &
\makecell[l]{Original Query: \textit{A dog \textcolor{red}{on} a leash \textcolor{red}{trying} to ride a skateboard.} \\ Paraphrased Query: \textit{A dog \textcolor{red}{restrained} by a leash \textcolor{red}{attempting} to ride a skateboard.}}
\tabularnewline
\midrule
{\includegraphics[width=\cellwidthsupp cm, height=\cellwidthsupp cm]{samples/000000099054.jpg}} &
\makecell[l]{Original Query: \textit{some \textcolor{red}{planes} on an \textcolor{red}{air port} run way} \\ Paraphrased Query: \textit{Some \textcolor{red}{aircraft are positioned} on a runway \textcolor{red}{at} an \textcolor{red}{airport}.}}
\tabularnewline
\midrule
{\includegraphics[width=\cellwidthsupp cm, height=\cellwidthsupp cm]{samples/000000425226.jpg}} &
\makecell[l]{Original Query: \textit{A man \textcolor{red}{kneeling down} in front of a \textcolor{red}{refrigerator}.} \\ Paraphrased Query: \textit{A man \textcolor{red}{on his knees} in front of a \textcolor{red}{fridge}.}}
\tabularnewline
\midrule
{\includegraphics[width=\cellwidthsupp cm, height=\cellwidthsupp cm]{samples/000000301563.jpg}} &
\makecell[l]{Original Query: \textit{A man \textcolor{red}{riding skis} down a \textcolor{red}{snow covered} slope.} \\ Paraphrased Query: \textit{A man \textcolor{red}{skiing} down a \textcolor{red}{snowy} slope.}}
\tabularnewline
\bottomrule
\end{tabular}
\vspace{0.1cm}
\caption{Paraphrased text-image retrieval dataset samples.}
\label{supp:fig:para_samples}
\end{figure*}

\clearpage

\subsection{Hyperparameters.}
We list the GPT-3 hyperparameters used for automatic paraphrase collection in Table \ref{supp:tab:hyperpara_gpt3}. 

\begin{table}[h]
\centering
\small
\footnotesize
\begin{tabular}{c|c}
\toprule
Hyperparameter & Value \\
\midrule
prompt & \textit{\makecell[l]{``Paraphrase the following sentence, using\\ as many synonyms as possible: \{Input\}''}} \\
engine & text-davinci-002 \\
temperature & 0.9 \\
max tokens & 32\\
\bottomrule
\end{tabular}
\caption{Hyperparameters for paraphrase collection by GPT-3.} \label{supp:tab:hyperpara_gpt3}
\end{table}

\section{Experiment}

\subsection{Hyperparameters.} 
We list the training hyperparameters used for all our models in Table \ref{supp:tab:hyperpara_clip}.

\begin{table}[h]
\centering
\small
\footnotesize
\begin{tabular}{c | c}
\toprule
Hyperparameter & Value \\
\midrule
batch size & 4000 (CC12M), 32760 (LAION-400M) \\
training epochs & 32 \\
warm-up iterations & 10000 \\
learning rate & 0.001 \\
AdamW weight decay & 0.1 \\
AdamW $\beta_1$ & 0.9 \\
AdamW $\beta_2$ & 0.999 (RN50), 0.98 (ViT-B/32) \\
AdamW $\epsilon$ & $10^{-8}$ (RN50), $10^{-6}$ (ViT-B/32) \\
\bottomrule
\end{tabular}
\caption{Hyperparameters for our training experiments.}
\label{supp:tab:hyperpara_clip}
\end{table}

\medskip

\subsection{Adaptation on CC12M.} 
The full ablation on the choice of image encoders (network architecture and initialization strategy) and adaptation methods is presented in Tables \ref{supp:tab:cc12m_1} and  \ref{supp:tab:cc12m_2}. 
\medskip

\subsection{Large-scale Adaptation on LAION-400M.} 
The performances for zero-shot classification and image retrieval  of different models  are compared in Table \ref{supp:tab:scaling}. The zero-shot classification accuracies of different models on 35 VTAB+ datasets are compared in Table \ref{supp:tab:vtab}. We present the top-10 retrieved images for the qualitative examples of Figure 4 in Figures \ref{supp:fig:qualitative1} and \ref{supp:fig:qualitative2}.
\medskip

\subsection{Query Expansion (QE) details.} 
For our evaluation of QE in Section \ref{sec:exp:qe}, we set the number of expanded queries as $K=3$. 


\begin{table*}[!t]
\begin{center}
\footnotesize
\vspace{-0.cm}
\begin{tabular}{ccc|ccccccccc}
\toprule
~ & ~ &  ~ &  \multicolumn{1}{c}{Classification}   &  \multicolumn{4}{c}{Text Retrieval}   &  \multicolumn{4}{c}{Image Retrieval} 
\tabularnewline
\midrule
  \multirow{2}{*}{\scriptsize{$\mathcal{V}$}}   & \multirow{2}{*}{Init. \scriptsize{$\mathcal{V}$}} &  \multirow{2}{*}{Method} & \multicolumn{1}{c}{ImageNet} & \multicolumn{2}{c}{Flickr30k} & \multicolumn{2}{c}{COCO}  & \multicolumn{2}{c}{Flickr30k} & \multicolumn{2}{c}{COCO} 
\tabularnewline
~ & ~ & ~ &   {Acc.}   & R@1 & R@5   & R@1 & R@5  & R@1 & R@5 & R@1 & R@5 
\tabularnewline
\midrule
 \multirow{10}{*}{{RN50}}  & \multirow{4}{*}{{Random}} &  \scriptsize{CLIP} & 33.6 & 47.5 & 75.8 & 19.4 & 42.0 & 43.0 & 71.8 & 18.9 & 41.5
\tabularnewline
~ & ~ & \scriptsize{$\mathcal{T}$ \textit{finetuned}} & \ccb 33.2 & \ccb 50.1 & \ccb 76.3 & \ccb 20.8 & \ccb  44.8 & \ccb 44.2 & \ccb 73.0 & \ccb 19.4 & \ccb 34.2 
\tabularnewline
~ & ~ & \scriptsize{$\mathcal{T}$ \textit{frozen}} &  20.1  &  31.1  &  60.9  & 12.4  & 31.8 & 28.1  &  55.4  &  11.9  &  27.6  \tabularnewline
~ & ~ & \scriptsize{$\mathcal{T}$ \textit{frozen + bottleneck}}  & \ccb  21.9 &  \ccb 32.7 &  \ccb 62.4 &  \ccb 13.0 &  \ccb 32.2 &  \ccb 29.6 &  \ccb 58.1 &  \ccb 12.3 & \ccb 29.5
\tabularnewline
~ & ~ & \scriptsize{$\mathcal{T}$ \textit{frozen + alignment}} & 33.7 &  54.7 &   80.5 &  23.9 & 47.7 &  49.7  & 76.5  & 20.5 &  44.5 
\tabularnewline
\cmidrule{2-12}
 ~  & \multirow{5}{*}{{ImageNet}} & \scriptsize{CLIP} &  \ccb 36.9 & \ccb  48.5 & \ccb 78.3 & \ccb 22.5 & \ccb 46.3  &  \ccb 48.2 &  \ccb 73.1 & \ccb  22.4 &  \ccb 45.8 
\tabularnewline
~ & ~ & \scriptsize{$\mathcal{T}$ \textit{finetuned}} & 37.3 &  51.6 & 79.5  & 24.2 & 47.3 & 48.8 &  75.7  & 22.7 & 46.2 
\tabularnewline
~ & ~ & \scriptsize{$\mathcal{T}$ \textit{frozen}} & \ccb  30.6 & \ccb 44.8  & \ccb 73.2   & \ccb  21.9 & \ccb 46.2 & \ccb 40.8  & \ccb 67.7   & \ccb 20.3  & \ccb  43.3  \tabularnewline
~ & ~ & \scriptsize{$\mathcal{T}$ \textit{frozen + bottleneck}} & 33.2 & 52.1 & 78.6 & 25.0 & 49.4 & 46.8 & 75.0 &  22.4 & 46.8 
\tabularnewline
~ & ~ & \scriptsize{$\mathcal{T}$ \textit{frozen + alignment}} &  \ccb 37.8  & \ccb 54.5 & \ccb 81.1 & \ccb 26.7  & \ccb 51.3 & \ccb 50.6 & \ccb 76.9 & \ccb  24.0 & \ccb 48.0 
\tabularnewline
\midrule
 \multirow{10}{*}{{ViT-B/32}}  & \multirow{4}{*}{{Random}} &  \scriptsize{CLIP} & 29.0 & 43.2 & 68.1 & 16.6 & 37.5 & 41.0 & 67.2 &  16.0 & 35.8   
\tabularnewline
~ & ~ & \scriptsize{$\mathcal{T}$ \textit{finetuned}} & \ccb 28.5 & \ccb 44.4 & \ccb 70.2 & \ccb 17.8 & \ccb 38.1  & \ccb 40.9  & \ccb 67.7 & \ccb 15.5 & \ccb 34.9 \tabularnewline
~ & ~ & \scriptsize{$\mathcal{T}$ \textit{frozen}} & 18.7  & 27.3  & 57.6 &  11.7  &  29.4  & 26.1  &  52.9   & 10.2  &   25.8 \tabularnewline
~ & ~ & \scriptsize{$\mathcal{T}$ \textit{frozen + bottleneck}} &  \ccb 18.8 &  \ccb 30.2 &  \ccb  57.8 &  \ccb 11.4 & \ccb 28.7 & \ccb 26.2 & \ccb 53.3 & \ccb 9.2 & \ccb 25.5 \tabularnewline
~ & ~ & \scriptsize{$\mathcal{T}$ \textit{frozen + alignment}} &  29.0 & 45.6  & 72.2 & 18.2 & 40.1 & 41.2 & 69.7 & 15.9 & 35.8  \tabularnewline
\cmidrule{2-12}
 ~  & \multirow{4}{*}{{ImageNet}} & \scriptsize{CLIP} & \ccb  36.8 & \ccb  52.3 &  \ccb 77.8 & \ccb  20.8 & \ccb  45.5 &  \ccb 49.7 & \ccb  75.1 &  \ccb 20.0 &  \ccb 42.0  \tabularnewline
~ & ~ & \scriptsize{$\mathcal{T}$ \textit{frozen}} &  30.2 & 42.9 & 71.6 &  16.5 & 38.6 & 39.2 & 66.9 &  16.0 & 38.6 \tabularnewline
~ & ~ & \scriptsize{$\mathcal{T}$ \textit{finetuned}} & \ccb 36.3 & \ccb 53.2 & \ccb 78.1 & \ccb 21.0 & \ccb 45.1  & \ccb 49.9 & \ccb 77.3 & \ccb 19.8 & \ccb 40.8 \tabularnewline 
~ & ~ & \scriptsize{$\mathcal{T}$ \textit{frozen + bottleneck}} & 30.7 & 42.3 & 70.9 & 17.2  &  38.8 & 39.2 & 67.3 & 15.1  &  36.3  \tabularnewline
~ & ~ & \scriptsize{$\mathcal{T}$ \textit{frozen + alignment}} & \ccb 37.5 & \ccb 54.6  & \ccb 80.7  & \ccb 21.8 & \ccb 46.4 & \ccb 51.9  & \ccb 80.0  & \ccb 20.6  & \ccb 45.9 \tabularnewline
\bottomrule
\end{tabular}
\end{center}
\caption{Performance of different models on zero-shot classification and image/text retrieval. All models are trained for 32 epochs on the CC12M dataset.}\label{supp:tab:cc12m_1}
\end{table*}

\begin{table*}[!t]
\begin{center}
\setlength{\tabcolsep}{1.4pt} 
\footnotesize
\vspace{-0.cm}
\begin{tabular}{ccc|cccccc}
\toprule
~ & ~ &  ~ &  \multicolumn{4}{c}{paraphrased text-image retrieval} & \multicolumn{2}{c}{semantic text similarity} 
\tabularnewline
\midrule
  \multirow{2}{*}{\scriptsize{$\mathcal{V}$}}   & \multirow{2}{*}{Init. \scriptsize{$\mathcal{V}$}} & \multirow{2}{*}{Method} & \multicolumn{2}{c}{COCO-P} & \multicolumn{2}{c}{COCO-PE}  & \multicolumn{2}{c}{STS-B} 
\tabularnewline
~ & ~ & ~ &  AO@10  & JS@10 &  AO@10  & JS@10  &  Pearson/\%  & Spearman/\%
\tabularnewline
\midrule
 \multirow{10}{*}{{RN50}}  & \multirow{5}{*}{{Random}} &  \scriptsize{CLIP} & 51.7  & 43.4 & 34.6  & 27.2 & 68.3  & 68.2  \tabularnewline
~ & ~ & \scriptsize{$\mathcal{T}$ \textit{finetuned}} & \ccb 53.0 & \ccb 44.9 & \ccb 35.1 & \ccb 28.1   & \ccb 71.0 &  \ccb 69.9 \tabularnewline 
~ & ~ & \scriptsize{$\mathcal{T}$ \textit{frozen}} &  68.6  &  60.6  &  54.2 & 45.4 & 82.8  &  83.9  \tabularnewline
~ & ~ & \scriptsize{$\mathcal{T}$ \textit{frozen + bottleneck}} &  \ccb  67.4 & \ccb  60.3  &  \ccb 53.9 &  \ccb  43.9  &  \ccb 73.6 &  \ccb 75.0  \tabularnewline 
~ & ~ & \scriptsize{$\mathcal{T}$ \textit{frozen + alignment}} & 69.0 & 60.9 &  55.3 & 46.3  & 76.7 & 78.8 \tabularnewline
\cmidrule{2-9}
 ~  & \multirow{5}{*}{{ImageNet}} & \scriptsize{CLIP} &   \ccb 53.7 &   \ccb 44.9 &  \ccb 35.7 &  \ccb 28.1 &  \ccb 68.4 & \ccb 68.5 \tabularnewline
~ & ~ & \scriptsize{$\mathcal{T}$ \textit{finetuned}} &  55.3 &  45.1 & 36.4 & 28.3  & 69.2 & 69.3  \tabularnewline
~ & ~ & \scriptsize{$\mathcal{T}$ \textit{frozen}} & \ccb 69.2  & \ccb 61.4  & \ccb  55.5  & \ccb  46.1 & \ccb 81.6 & \ccb 83.1  \tabularnewline
~ & ~ & \scriptsize{$\mathcal{T}$ \textit{frozen + bottleneck}}  & 68.6 &  60.7  & 54.5 & 45.3  & 73.6 & 75.8 \tabularnewline
~ & ~ & \scriptsize{$\mathcal{T}$ \textit{frozen + alignment}} & \ccb 69.8 & \ccb 61.2 & \ccb 55.3 & \ccb 46.0 &  \ccb 76.7 & \ccb  78.7 \tabularnewline
\midrule
 \multirow{10}{*}{{ViT-B/32}}  & \multirow{5 }{*}{{Random}} &  \scriptsize{CLIP}  & 50.9  &  42.5  & 33.8 & 26.4  & 71.8 & 71.6 \tabularnewline
~ & ~ & \scriptsize{$\mathcal{T}$ \textit{finetuned}} & \ccb 51.9 & \ccb 44.9 & \ccb 34.0 & \ccb 26.7 & \ccb 72.4 & \ccb 72.9  \tabularnewline
~ & ~ & \scriptsize{$\mathcal{T}$ \textit{frozen}} & 65.6 &  56.9  &  52.0 &  42.9  & 82.1 & 83.3  \tabularnewline
~ & ~ & \scriptsize{$\mathcal{T}$ \textit{frozen + bottleneck}} &  \ccb 65.5 &  \ccb  57.0 & \ccb  52.7 &  \ccb 43.1  &  \ccb 82.2 &  \ccb 83.4  \tabularnewline
~ & ~ & \scriptsize{$\mathcal{T}$ \textit{frozen + alignment}} & 68.3 & 60.2 & 55.2 & 46.1  & 79.0 & 79.6 \tabularnewline
\cmidrule{2-9}
 ~  & \multirow{5}{*}{{ImageNet}} & \scriptsize{CLIP} &  \ccb 53.2  & \ccb  44.6 & \ccb 35.5 & \ccb  28.0 & \ccb 70.0 & \ccb 69.7  \tabularnewline
~ & ~ & \scriptsize{$\mathcal{T}$ \textit{finetuned}} &  53.9 & 44.5 & 35.6 & 28.4  & 72.1 & 72.9 \tabularnewline
~ & ~ & \scriptsize{$\mathcal{T}$ \textit{frozen}} & \ccb  66.6 & \ccb 58.6   & \ccb  53.4 & \ccb  43.9 & \ccb 81.1 & \ccb 82.9 \tabularnewline
~ & ~ & \scriptsize{$\mathcal{T}$ \textit{frozen + bottleneck}} & 66.8 & 58.1  & 52.8 & 43.4   & 81.7 & 83.2 \tabularnewline
~ & ~ & \scriptsize{$\mathcal{T}$ \textit{frozen + alignment}} & \ccb 68.8 & \ccb 60.7 & \ccb 55.1 & \ccb 45.9 & \ccb 78.9 & \ccb 80.0 \tabularnewline
\bottomrule
\end{tabular}
\end{center}
\caption{Performance of different models on paraphrased text-image retrieval and semantic text similarity. All models are trained for 32 epochs on the CC12M dataset.}\label{supp:tab:cc12m_2}
\end{table*}

\begin{table*}[h]
\centering
\footnotesize
\scriptsize
\begin{tabular}{*l c | c c c c c c | c c c c c | c c c c c}
\toprule
& &
\multicolumn{3}{c}{\scriptsize classification} & 
\multicolumn{6}{c}{\scriptsize image retrieval} \\
 &  & 
\multicolumn{2}{c}{\scriptsize IN-1K} &  
\multicolumn{1}{c}{\scriptsize VTAB+} & 
\multicolumn{3}{c}{\scriptsize Flickr30k} & \multicolumn{3}{c}{\scriptsize COCO} \\
model & data &
\multicolumn{1}{c}{\scriptsize Top-1 Acc.} &
\multicolumn{1}{c}{\scriptsize Top-5 Acc.} &
\multicolumn{1}{c}{\scriptsize Top-1 Acc.} &
\multicolumn{1}{c}{\scriptsize R@1} & 
\multicolumn{1}{c}{\scriptsize R@5} & 
\multicolumn{1}{c}{\scriptsize R@10} & 
\multicolumn{1}{c}{\scriptsize R@1} & 
\multicolumn{1}{c}{\scriptsize R@5} & 
\multicolumn{1}{c}{\scriptsize R@10} \\
\midrule
CLIP\cite{radford2021learning} & WIT-400M &
\better{63.3} & \better{88.8} & 
45.5 &
58.8 & 83.4 & 89.8 &
30.4 & 56.0 & 66.9 \\
OpenCLIP \cite{cherti2022reproducible} & LAION-400M &
62.9 & 87.7 & 
45.8 &
\better{61.7} & \better{85.5} & \better{90.9} & 
\better{35.3} & \better{60.9} & \better{71.6} \\
\hline
Ours & LAION-400M  &
60.5 & 86.0 & 
\better{46.6} &
59.9 & 84.0 & 90.7 & 
33.9 & 60.0 & 70.6 \\
\bottomrule
\end{tabular}

\caption{{Large scale adaptation on LAION-400M. All models use the ViT-B/32 architecture as the visual backbone and are trained for 32 epochs on the specified dataset.}}
\label{supp:tab:scaling}
\end{table*}

\begin{table*}[h]
\centering
\scriptsize
\setlength{\tabcolsep}{3pt}
\begin{tabular}{*l | c | c c c c c c | c c c c c| c c c c c| c}
\toprule
& &
\multicolumn{6}{c}{\scriptsize Character-level} & 
\multicolumn{5}{c}{\scriptsize Word-level}  & 
\multicolumn{5}{c}{\scriptsize Sentence-level} & \\
 & Clean & 
{K} & 
{OCR} & 
{CI} & 
{CR} & 
{CS} & 
{CD} & 
{SR} & 
{WI} & 
{WS} & 
{WD} & 
{IP} & 
{P} & 
{A} & 
{F} & 
{C} & 
{BT} & 
 Avg.
 \\
\midrule
\multirow{2}{*}{CLIP}  &  361.8  
 &  235.0  &  294.4    &  253.3   &  236.7   &  247.1   &  249.6 
 &  313.3   &  330.1   &  319.2   &  332.4   &  344.2  
 &  348.7   &  359.6   &  357.9   &  357.2   &  350.9  
 &  308.1 
\\
 & 
 & $\downarrow$35.1\%  & $\downarrow$18.6\%  & $\downarrow$30.0\%  & $\downarrow$34.6\%  & $\downarrow$31.7\%  & $\downarrow$31.0\% 
 & $\downarrow$13.4 \%  & $\downarrow$8.8\%  & $\downarrow$11.8\%  & $\downarrow$8.1\%  & $\downarrow$4.9\% 
 & $\downarrow$3.6\%  & $\downarrow$0.6\%  & $\downarrow$1.1\%  & $\downarrow$1.3\%  & $\downarrow$3.0\%  
 & $\downarrow$14.8\% 
\\
\midrule
\multirow{2}{*}{OpenCLIP}  &  381.8
 & 249.5 & 315.1   & 269.2  & 251.8  & 265.1  & 265.4 
 & 331.4  & 357.5  & 355.6  & 358.4  & 370.1 
 & 370.7  & 381.5  & 378.7  & 377.6  & 371.6 
 & 329.3 
\\
 & 
 & $\downarrow$34.6\%  & $\downarrow$17.5\%  & $\downarrow$29.5\%  & $\downarrow$34.0\%  & $\downarrow$30.6\%  & $\downarrow$30.5\% 
 & $\downarrow$13.2\%  & $\downarrow$6.4\%  & $\downarrow$6.9\%  & $\downarrow$6.1\%  & $\downarrow$3.1\% 
 & $\downarrow$2.9\%  & $\downarrow$0.1\%  & $\downarrow$0.8\%  & $\downarrow$1.1\%  & $\downarrow$2.7\%  
 & $\downarrow$13.7\% 
\\
\midrule
\multirow{2}{*}{Ours}  & 371.6  
 & 277.1  &   327.2   &  310.6  &  278.4   &  293.9   &  290.1 
 &  334.3   &  353.6   &  355.9   &  352.4  &  361.6 
 &  362.8    &  370.6   &  368.4   &  368.6   &  358.6 
 &  335.3 
\\
 & 
 &  $\downarrow$25.4\%   &  $\downarrow$11.9\%    & $\downarrow$16.4\%   &  $\downarrow$25.1\%   &  $\downarrow$20.9\%  &  $\downarrow$21.9\% 
 &  $\downarrow$10.0\%   &  $\downarrow$4.8\%  &  $\downarrow$4.2\%   & $\downarrow$5.1\%   &  $\downarrow$2.7\% 
 &  $\downarrow$2.3\%   &  $\downarrow$0.3\%   &  $\downarrow$0.9\%   &  $\downarrow$0.8\%   &  $\downarrow$3.5\% 
 &  $\downarrow$9.8\% 
\\

\bottomrule
\end{tabular}

\caption{Evaluation of robustness to three different levels of text perturbation on COCO-TP \cite{qiu2022multimodal}. The types of perturbation are: a) \textbf{Character-level}: keyboard (K), OCR, character insert (CI), character replace (CR), character swap (CS), character delete (CD); b) \textbf{Word-level}: synonym replacement (SR), word insertion (WR), word swap (WS), word deletion (WD), and insert punctuation (IP); c) \textbf{Sentence-level}: formal (F), casual (C), passive (P), active (A), back translation (BT).}
\label{supp:COCO-TP}

\end{table*}

\begin{table*}[h]
\centering
\footnotesize
\begin{tabular}{ c |ccc}
\toprule
 & CLIP \cite{radford2021learning} &  OpenCLIP \cite{cherti2022reproducible} & Ours \\
data & WIT-400M & LAION-400M  & LAION-400M \\
\midrule
ImageNet & \better{63.3} & 62.9 & 60.5 \\
ImageNet-v2 & \better{55.9} & 55.1 & 52.5 \\
ImageNet-R & 69.2 & \better{73.5} & 71.0 \\
ImageNet-Sketch & 42.3 & \better{49.4} & 46.6 \\
ObjectNet & \better{44.2} & 43.9 & 42.8 \\
ImageNet-A & \better{31.4} & 21.7 & 21.1 \\
CIFAR-10 & 89.8 & 90.8 & \better{92.4} \\
CIFAR-100 & 64.3 & 70.3  & \better{71.8} \\
MNIST & 48.7 & 37.6 & \better{61.6} \\
Flowers102 & 66.4 & \better{68.0} & 56.5 \\
Cars & 59.7 & \better{79.3} & 78.3 \\
SVHN & 13.5 & 27.7 & \better{38.3} \\
FER2013 & 41.3 & \better{43.0} & 42.9  \\
RenderedSST2 & \better{58.6} & 52.4 & 56.7 \\
Pets & \better{87.3} & 86.9 & 83.2 \\
Caltech-101 & 81.6 & \better{83.2} & 82.9 \\
VOC2007-Cl & \better{76.4} & 75.8 & 75.5 \\
SUN397 & 62.3 & \better{67.0} & 66.0 \\
FGVC Aircraft & \better{19.6} & 16.6 & 12.0 \\
Country211 & \better{17.2} & 14.8 & 13.9 \\
DTD & 44.3 & {54.6} & \better{56.0} \\
GTSRB & 32.6 & 42.0 & \better{44.1} \\
STL10 & \better{97.1} & 95.6 & 96.5 \\
Retino & 44.8 & 25.2 & \better{63.2} \\
EuroSAT & 50.5 & \better{51.5} & 42.2 \\
RESISC45 & 53.7 & \better{54.6} & 52.7 \\
PCAM & \better{62.3} & 55.8 & 61.4 \\
CLEVR Counts & \better{23.2} & 16.2 & {16.3}  \\
CLEVR Dist & 23.4 & \better{23.9} & 20.0 \\
DSPRITES Orient & \better{2.5} & 1.9 & 2.2 \\
DSPRITES Pos & \better{3.4} & 3.0 & 3.2 \\
SmallNORB Elv & \better{12.8} & 4.5 & 11.3 \\
SmallNORB Azim & 6.1 & \better{9.9} & 4.9 \\
DMLAB & \better{19.3} & 17.3  & 17.3 \\
KITTI Dist & 27.4 & \better{29.1} & 17.6 \\
\midrule
VTAB+ (Avg.) & 45.5 & 45.8 & \better{46.6} \\
\bottomrule
\end{tabular}
\caption{Zero-shot classification accuracies on VTAB+ 35 tasks. All models use the ViT-B/32 architecture as the visual backbone and are trained for 32 epochs on the specified dataset.}
\label{supp:tab:vtab}
\end{table*}

\clearpage

\begin{figure*}[!t]
\begin{center}
\setlength{\tabcolsep}{3pt} 
\renewcommand{\arraystretch}{0.0} 
\begin{tabular}{|c|cccccccccc|}
\toprule
 & \multicolumn{10}{|c|}{$q_1$: \textit{``A road sign that indicates route \textcolor{red}{sixty six}. ''} (top row)} \tabularnewline
\cmidrule{2-11}
 & \multicolumn{10}{|c|}{$p_1$: \textit{``A road sign that indicates Route \textcolor{red}{66}. ''} (bottom row)} \tabularnewline
\midrule
\multirow{2}{*}{CLIP \cite{radford2021learning}}  &
\raisebox{-0.\height}{\highlight{\includegraphics[width=\cellwidth cm, height=\cellwidth cm]{samples/resized/3024/000000101022.jpg}}}&
\raisebox{-0.\height}{\highlight{\includegraphics[width=\cellwidth cm, height=\cellwidth cm]{samples/resized/3024/000000108536.jpg}}}&
\raisebox{-0.\height}{\includegraphics[width=\cellwidth cm, height=\cellwidth cm]{samples/resized/3024/000000211439.jpg}}&
\raisebox{-0.\height}{\includegraphics[width=\cellwidth cm, height=\cellwidth cm]{samples/resized/3024/000000325690.jpg}}&
\raisebox{-0.\height}{\includegraphics[width=\cellwidth cm, height=\cellwidth cm]{samples/resized/3024/000000246973.jpg}}&
\raisebox{-0.\height}{\includegraphics[width=\cellwidth cm, height=\cellwidth cm]{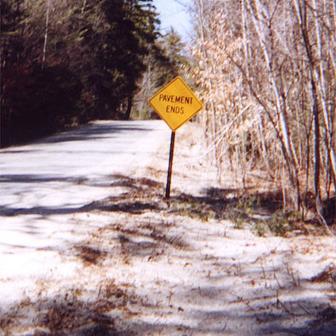}}&
\raisebox{-0.\height}{\includegraphics[width=\cellwidth cm, height=\cellwidth cm]{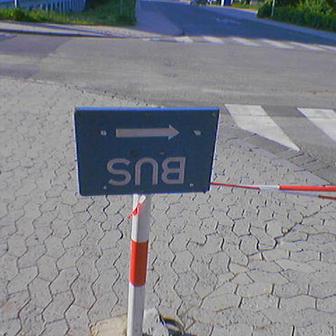}}&
\raisebox{-0.\height}{\includegraphics[width=\cellwidth cm, height=\cellwidth cm]{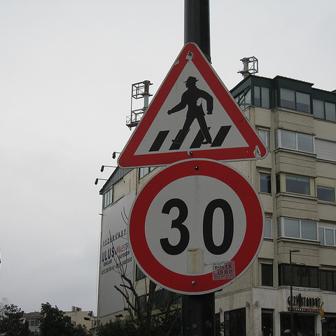}}&
\raisebox{-0.\height}{\includegraphics[width=\cellwidth cm, height=\cellwidth cm]{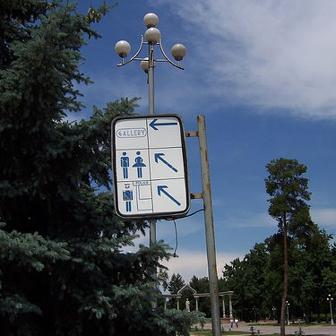}}&
\raisebox{-0.\height}{\highlight{\includegraphics[width=\cellwidth cm, height=\cellwidth cm]{samples/resized/3024/000000396415.jpg}}}
\tabularnewline
\cmidrule{2-11}
 ~  &
\raisebox{-0.\height}{\includegraphics[width=\cellwidth cm, height=\cellwidth cm]{samples/resized/3024/000000059250.jpg}}&
\raisebox{-0.\height}{\includegraphics[width=\cellwidth cm, height=\cellwidth cm]{samples/resized/3024/000000007592.jpg}}&
\raisebox{-0.\height}{\highlight{\includegraphics[width=\cellwidth cm, height=\cellwidth cm]{samples/resized/3024/000000108536.jpg}}}&
\raisebox{-0.\height}{\highlight{\includegraphics[width=\cellwidth cm, height=\cellwidth cm]{samples/resized/3024/000000396415.jpg}}}&
\raisebox{-0.\height}{\highlight{\includegraphics[width=\cellwidth cm, height=\cellwidth cm]{samples/resized/3024/000000101022.jpg}}}&
\raisebox{-0.\height}{\includegraphics[width=\cellwidth cm, height=\cellwidth cm]{samples/resized/3024/000000302945.jpg}}&
\raisebox{-0.\height}{\includegraphics[width=\cellwidth cm, height=\cellwidth cm]{samples/resized/3024/000000432370.jpg}}&
\raisebox{-0.\height}{\includegraphics[width=\cellwidth cm, height=\cellwidth cm]{samples/resized/3024/000000189735.jpg}}&
\raisebox{-0.\height}{\includegraphics[width=\cellwidth cm, height=\cellwidth cm]{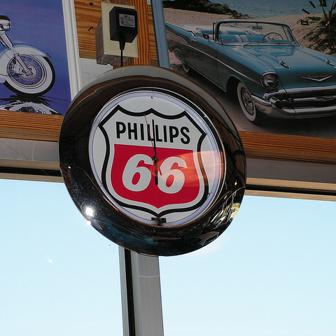}}&
\raisebox{-0.\height}{\includegraphics[width=\cellwidth cm, height=\cellwidth cm]{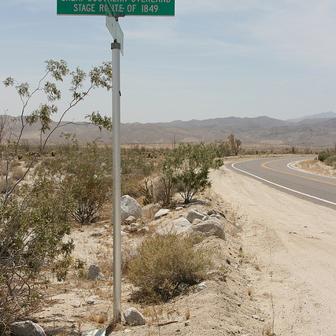}}
\tabularnewline
\midrule
\multirow{2}{*}{Ours}   &
\raisebox{-0.\height}{\highlight{\includegraphics[width=\cellwidth cm, height=\cellwidth cm]{samples/resized/3024/000000029287.jpg}}}& 
\raisebox{-0.\height}{\highlight{\includegraphics[width=\cellwidth cm, height=\cellwidth cm]{samples/resized/3024/000000189735.jpg}}}&
\raisebox{-0.\height}{\highlight{\includegraphics[width=\cellwidth cm, height=\cellwidth cm]{samples/resized/3024/000000521682.jpg}}}&
\raisebox{-0.\height}{\highlight{\includegraphics[width=\cellwidth cm, height=\cellwidth cm]{samples/resized/3024/000000220932.jpg}}}&
\raisebox{-0.\height}{\includegraphics[width=\cellwidth cm, height=\cellwidth cm]{samples/resized/3024/000000101022.jpg}}&
\raisebox{-0.\height}{\includegraphics[width=\cellwidth cm, height=\cellwidth cm]{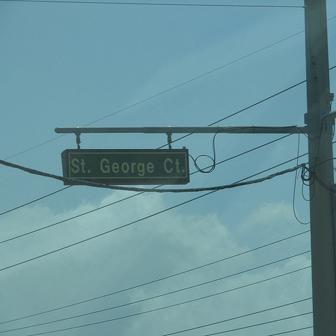}}&
\raisebox{-0.\height}{\highlight{\includegraphics[width=\cellwidth cm, height=\cellwidth cm]{samples/resized/3024/000000059250.jpg}}}&
\raisebox{-0.\height}{\highlight{\includegraphics[width=\cellwidth cm, height=\cellwidth cm]{samples/resized/3024/000000432370.jpg}}}&
\raisebox{-0.\height}{\includegraphics[width=\cellwidth cm, height=\cellwidth cm]{samples/resized/3024/000000309322.jpg}}&
\raisebox{-0.\height}{\includegraphics[width=\cellwidth cm, height=\cellwidth cm]{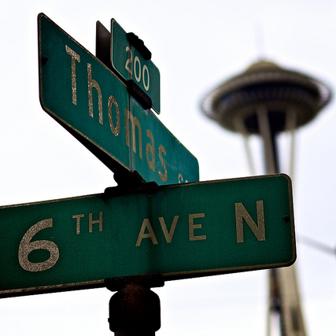}}
\tabularnewline
\cmidrule{2-11}
 ~  &
\raisebox{-0.\height}{\highlight{\includegraphics[width=\cellwidth cm, height=\cellwidth cm]{samples/resized/3024/000000059250.jpg}}}&
\raisebox{-0.\height}{\highlight{\includegraphics[width=\cellwidth cm, height=\cellwidth cm]{samples/resized/3024/000000432370.jpg}}}&
\raisebox{-0.\height}{\highlight{\includegraphics[width=\cellwidth cm, height=\cellwidth cm]{samples/resized/3024/000000189735.jpg}}}&
\raisebox{-0.\height}{\highlight{\includegraphics[width=\cellwidth cm, height=\cellwidth cm]{samples/resized/3024/000000029287.jpg}}}&
\raisebox{-0.\height}{\includegraphics[width=\cellwidth cm, height=\cellwidth cm]{samples/resized/3024/000000302945.jpg}}&
\raisebox{-0.\height}{\highlight{\includegraphics[width=\cellwidth cm, height=\cellwidth cm]{samples/resized/3024/000000521682.jpg}}}&
\raisebox{-0.\height}{\includegraphics[width=\cellwidth cm, height=\cellwidth cm]{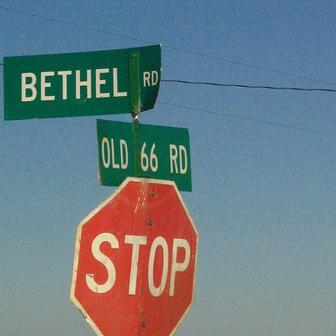}}&
\raisebox{-0.\height}{\includegraphics[width=\cellwidth cm, height=\cellwidth cm]{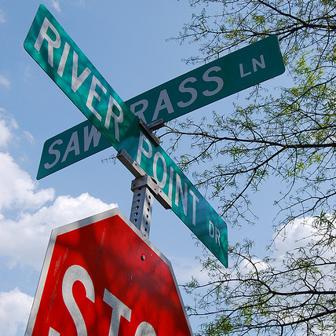}}&
\raisebox{-0.\height}{\highlight{\includegraphics[width=\cellwidth cm, height=\cellwidth cm]{samples/resized/3024/000000220932.jpg}}}&
\raisebox{-0.\height}{\includegraphics[width=\cellwidth cm, height=\cellwidth cm]{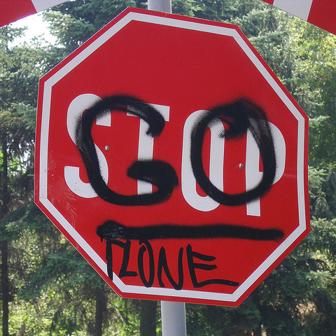}}
\tabularnewline
\bottomrule
\toprule
 & \multicolumn{10}{|c|}{$q_2$: \textit{``\textcolor{red}{An} open \textcolor{red}{refrigerator filled with lots} of food.''} (top row)} \tabularnewline
\cmidrule{2-11}
 & \multicolumn{10}{|c|}{$p_2$: \textit{``\textcolor{red}{A fridge that is} open \textcolor{red}{and has a lot} of food \textcolor{red}{inside}.''} (bottom row)} \tabularnewline
\midrule
 \multirow{2}{*}{CLIP \cite{radford2021learning}} &
\raisebox{-0.\height}{\highlight{\includegraphics[width=\cellwidth cm, height=\cellwidth cm]{samples/resized/refrigerator/000000455974.jpg}}}&
\raisebox{-0.\height}{\includegraphics[width=\cellwidth cm, height=\cellwidth cm]{samples/resized/refrigerator/000000044267.jpg}}&
\raisebox{-0.\height}{\highlight{\includegraphics[width=\cellwidth cm, height=\cellwidth cm]{samples/resized/refrigerator/000000234328.jpg}}}&
\raisebox{-0.\height}{\includegraphics[width=\cellwidth cm, height=\cellwidth cm]{samples/resized/refrigerator/000000552520.jpg}}&
\raisebox{-0.\height}{\includegraphics[width=\cellwidth cm, height=\cellwidth cm]{samples/resized/refrigerator/000000555156.jpg}}&
\raisebox{-0.\height}{\highlight{\includegraphics[width=\cellwidth cm, height=\cellwidth cm]{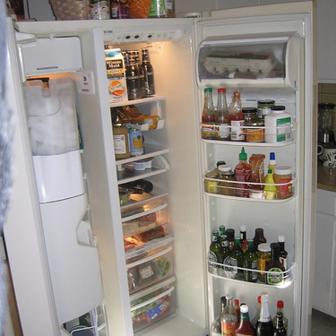}}}&
\raisebox{-0.\height}{\highlight{\includegraphics[width=\cellwidth cm, height=\cellwidth cm]{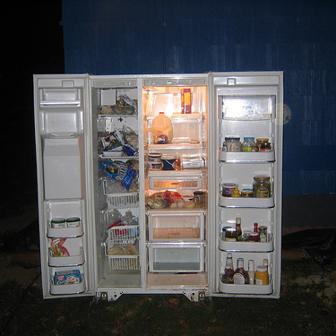}}}&
\raisebox{-0.\height}{\includegraphics[width=\cellwidth cm, height=\cellwidth cm]{samples/resized/refrigerator/000000541856.jpg}}&
\raisebox{-0.\height}{\includegraphics[width=\cellwidth cm, height=\cellwidth cm]{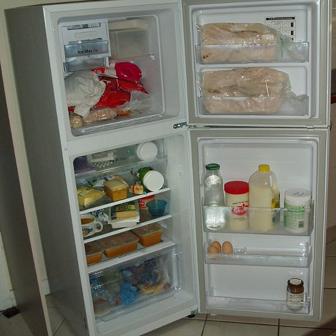}}&
\raisebox{-0.\height}{\includegraphics[width=\cellwidth cm, height=\cellwidth cm]{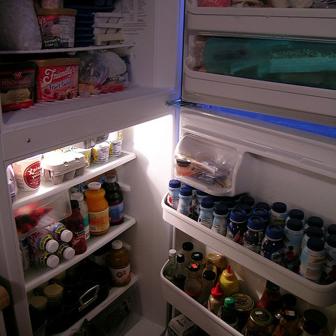}}
\tabularnewline
\cmidrule{2-11}
 ~  &
\raisebox{-0.\height}{\includegraphics[width=\cellwidth cm, height=\cellwidth cm]{samples/resized/refrigerator/000000163125.jpg}}&
\raisebox{-0.\height}{\highlight{\includegraphics[width=\cellwidth cm, height=\cellwidth cm]{samples/resized/refrigerator/000000234328.jpg}}}&
\raisebox{-0.\height}{\includegraphics[width=\cellwidth cm, height=\cellwidth cm]{samples/resized/refrigerator/000000396120.jpg}}&
\raisebox{-0.\height}{\includegraphics[width=\cellwidth cm, height=\cellwidth cm]{samples/resized/refrigerator/000000489497.jpg}}&
\raisebox{-0.\height}{\includegraphics[width=\cellwidth cm, height=\cellwidth cm]{samples/resized/refrigerator/000000024787.jpg}}&
\raisebox{-0.\height}{\highlight{\includegraphics[width=\cellwidth cm, height=\cellwidth cm]{samples/resized/refrigerator/000000165162.jpg}}}&
\raisebox{-0.\height}{\highlight{\includegraphics[width=\cellwidth cm, height=\cellwidth cm]{samples/resized/refrigerator/000000341218.jpg}}}&
\raisebox{-0.\height}{\highlight{\includegraphics[width=\cellwidth cm, height=\cellwidth cm]{samples/resized/refrigerator/000000455974.jpg}}}&
\raisebox{-0.\height}{\includegraphics[width=\cellwidth cm, height=\cellwidth cm]{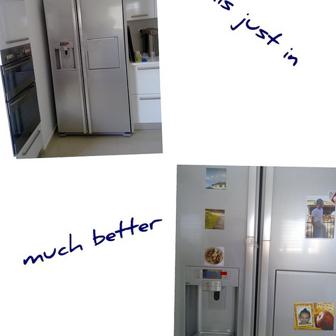}}&
\raisebox{-0.\height}{\includegraphics[width=\cellwidth cm, height=\cellwidth cm]{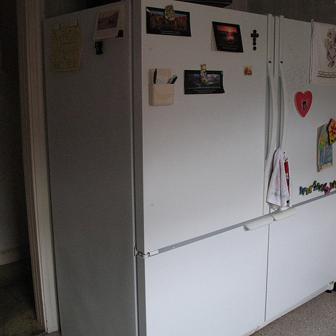}}
\tabularnewline
\midrule
\multirow{2}{*}{Ours}   &
\raisebox{-0.\height}{\highlight{\includegraphics[width=\cellwidth cm, height=\cellwidth cm]{samples/resized/refrigerator/000000064704.jpg}}}&
\raisebox{-0.\height}{\highlight{\includegraphics[width=\cellwidth cm, height=\cellwidth cm]{samples/resized/refrigerator/000000541856.jpg}}}&
\raisebox{-0.\height}{\highlight{\includegraphics[width=\cellwidth cm, height=\cellwidth cm]{samples/resized/refrigerator/000000555156.jpg}}}&
\raisebox{-0.\height}{\highlight{\includegraphics[width=\cellwidth cm, height=\cellwidth cm]{samples/resized/refrigerator/000000017909.jpg}}}&
\raisebox{-0.\height}{\includegraphics[width=\cellwidth cm, height=\cellwidth cm]{samples/resized/refrigerator/000000042427.jpg}}&
\raisebox{-0.\height}{\includegraphics[width=\cellwidth cm, height=\cellwidth cm]{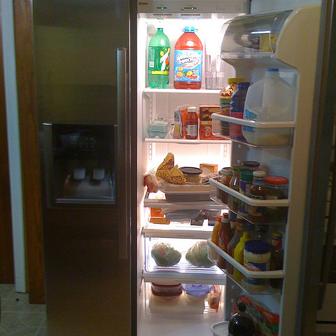}}&
\raisebox{-0.\height}{\highlight{\includegraphics[width=\cellwidth cm, height=\cellwidth cm]{samples/resized/refrigerator/000000150300.jpg}}}&
\raisebox{-0.\height}{\highlight{\includegraphics[width=\cellwidth cm, height=\cellwidth cm]{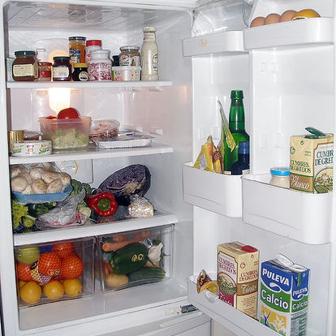}}}&
\raisebox{-0.\height}{\includegraphics[width=\cellwidth cm, height=\cellwidth cm]{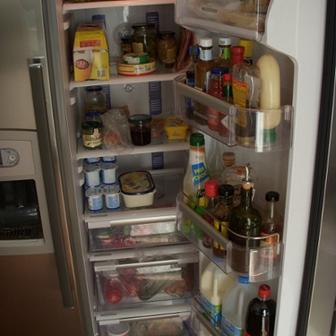}}&
\raisebox{-0.\height}{\includegraphics[width=\cellwidth cm, height=\cellwidth cm]{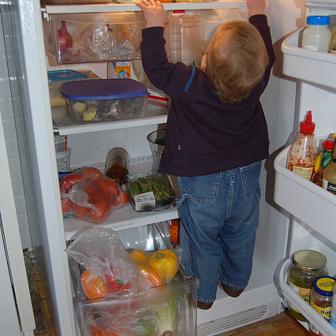}}
\tabularnewline
\cmidrule{2-11}
 ~  &
\raisebox{-0.\height}{\highlight{\includegraphics[width=\cellwidth cm, height=\cellwidth cm]{samples/resized/refrigerator/000000064704.jpg}}}&
\raisebox{-0.\height}{\highlight{\includegraphics[width=\cellwidth cm, height=\cellwidth cm]{samples/resized/refrigerator/000000555156.jpg}}}&
\raisebox{-0.\height}{\highlight{\includegraphics[width=\cellwidth cm, height=\cellwidth cm]{samples/resized/refrigerator/000000541856.jpg}}}&
\raisebox{-0.\height}{\highlight{\includegraphics[width=\cellwidth cm, height=\cellwidth cm]{samples/resized/refrigerator/000000150300.jpg}}}&
\raisebox{-0.\height}{\includegraphics[width=\cellwidth cm, height=\cellwidth cm]{samples/resized/refrigerator/000000024234.jpg}}&
\raisebox{-0.\height}{\highlight{\includegraphics[width=\cellwidth cm, height=\cellwidth cm]{samples/resized/refrigerator/000000017909.jpg}}}&
\raisebox{-0.\height}{\includegraphics[width=\cellwidth cm, height=\cellwidth cm]{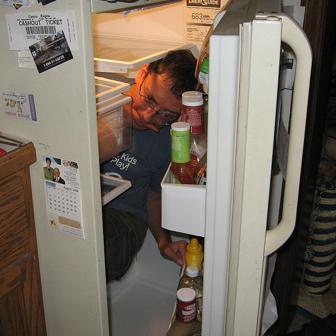}}&
\raisebox{-0.\height}{\includegraphics[width=\cellwidth cm, height=\cellwidth cm]{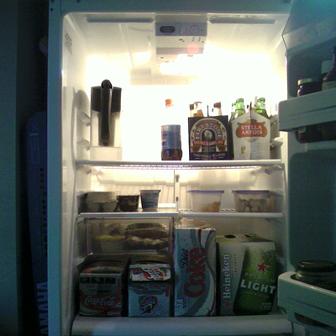}}&
\raisebox{-0.\height}{\includegraphics[width=\cellwidth cm, height=\cellwidth cm]{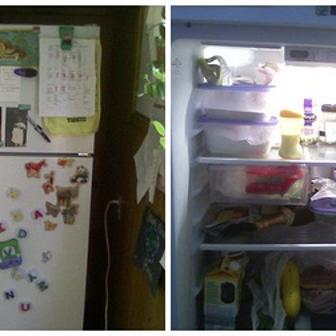}}&
\raisebox{-0.\height}{\highlight{\includegraphics[width=\cellwidth cm, height=\cellwidth cm]{samples/resized/refrigerator/000000034915.jpg}}}
\tabularnewline
\bottomrule
\toprule
 & \multicolumn{10}{|c|}{$q_3$: \textit{``A cat \textcolor{red}{sitting on the back of} a sofa \textcolor{red}{looking} out a window.''} (top row)} \tabularnewline
\cmidrule{2-11}
 & \multicolumn{10}{|c|}{$p_3$: \textit{``A cat \textcolor{red}{perched atop} a sofa \textcolor{red}{gazing} out a window. ''} (bottom row)} \tabularnewline
\midrule
 \multirow{2}{*}{CLIP \cite{radford2021learning}} &
\raisebox{-0.\height}{\includegraphics[width=\cellwidth cm, height=\cellwidth cm]{samples/resized/3819/000000109483.jpg}}&
\raisebox{-0.\height}{\includegraphics[width=\cellwidth cm, height=\cellwidth cm]{samples/resized/3819/000000483928.jpg}}&
\raisebox{-0.\height}{\highlight{\includegraphics[width=\cellwidth cm, height=\cellwidth cm]{samples/resized/3819/000000563869.jpg}}}&
\raisebox{-0.\height}{\highlight{\includegraphics[width=\cellwidth cm, height=\cellwidth cm]{samples/resized/3819/000000354630.jpg}}}&
\raisebox{-0.\height}{\includegraphics[width=\cellwidth cm, height=\cellwidth cm]{samples/resized/3819/000000036962.jpg}}&
\raisebox{-0.\height}{\includegraphics[width=\cellwidth cm, height=\cellwidth cm]{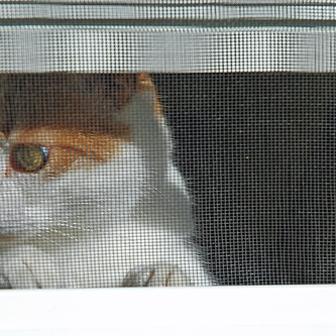}}&
\raisebox{-0.\height}{\highlight{\includegraphics[width=\cellwidth cm, height=\cellwidth cm]{samples/resized/3819/000000393995.jpg}}}&
\raisebox{-0.\height}{\highlight{\includegraphics[width=\cellwidth cm, height=\cellwidth cm]{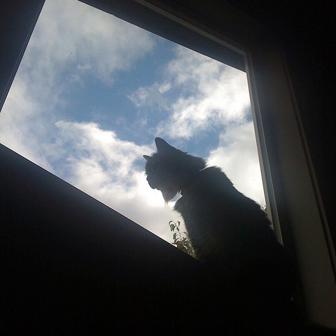}}}&
\raisebox{-0.\height}{\includegraphics[width=\cellwidth cm, height=\cellwidth cm]{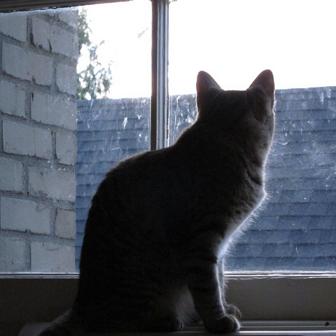}}&
\raisebox{-0.\height}{\highlight{\includegraphics[width=\cellwidth cm, height=\cellwidth cm]{samples/resized/3819/000000125100.jpg}}}
\tabularnewline
\cmidrule{2-11}
 ~  &
\raisebox{-0.\height}{\highlight{\includegraphics[width=\cellwidth cm, height=\cellwidth cm]{samples/resized/3819/000000563869.jpg}}}&
\raisebox{-0.\height}{\highlight{\includegraphics[width=\cellwidth cm, height=\cellwidth cm]{samples/resized/3819/000000125100.jpg}}}&
\raisebox{-0.\height}{\highlight{\includegraphics[width=\cellwidth cm, height=\cellwidth cm]{samples/resized/3819/000000393995.jpg}}}&
\raisebox{-0.\height}{\includegraphics[width=\cellwidth cm, height=\cellwidth cm]{samples/resized/3819/000000376472.jpg}}&
\raisebox{-0.\height}{\includegraphics[width=\cellwidth cm, height=\cellwidth cm]{samples/resized/3819/000000386291.jpg}}&
\raisebox{-0.\height}{\highlight{\includegraphics[width=\cellwidth cm, height=\cellwidth cm]{samples/resized/3819/000000354630.jpg}}}&
\raisebox{-0.\height}{\highlight{\includegraphics[width=\cellwidth cm, height=\cellwidth cm]{samples/resized/3819/000000056857.jpg}}}&
\raisebox{-0.\height}{\includegraphics[width=\cellwidth cm, height=\cellwidth cm]{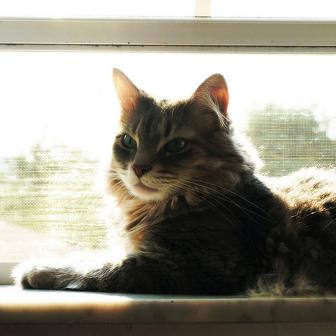}}&
\raisebox{-0.\height}{\includegraphics[width=\cellwidth cm, height=\cellwidth cm]{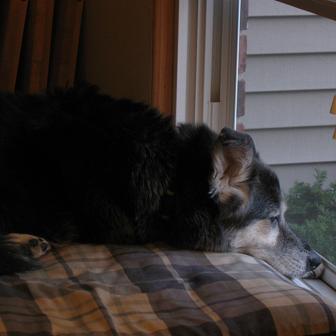}}&
\raisebox{-0.\height}{\includegraphics[width=\cellwidth cm, height=\cellwidth cm]{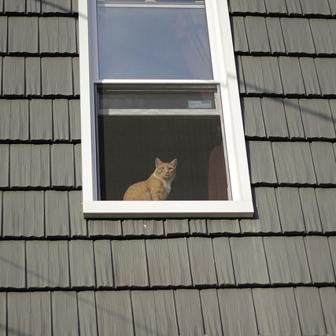}}
\tabularnewline
\midrule
\multirow{2}{*}{Ours}   &
\raisebox{-0.\height}{\highlight{\includegraphics[width=\cellwidth cm, height=\cellwidth cm]{samples/resized/3819/000000070353.jpg}}}&
\raisebox{-0.\height}{\highlight{\includegraphics[width=\cellwidth cm, height=\cellwidth cm]{samples/resized/3819/000000226220.jpg}}}&
\raisebox{-0.\height}{\highlight{\includegraphics[width=\cellwidth cm, height=\cellwidth cm]{samples/resized/3819/000000342078.jpg}}}&
\raisebox{-0.\height}{\highlight{\includegraphics[width=\cellwidth cm, height=\cellwidth cm]{samples/resized/3819/000000237419.jpg}}}&
\raisebox{-0.\height}{\includegraphics[width=\cellwidth cm, height=\cellwidth cm]{samples/resized/3819/000000539584.jpg}}&
\raisebox{-0.\height}{\highlight{\includegraphics[width=\cellwidth cm, height=\cellwidth cm]{samples/resized/3819/000000132578.jpg}}}&
\raisebox{-0.\height}{\includegraphics[width=\cellwidth cm, height=\cellwidth cm]{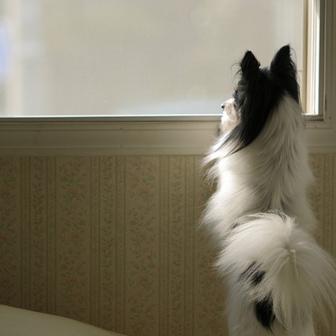}}&
\raisebox{-0.\height}{\highlight{\includegraphics[width=\cellwidth cm, height=\cellwidth cm]{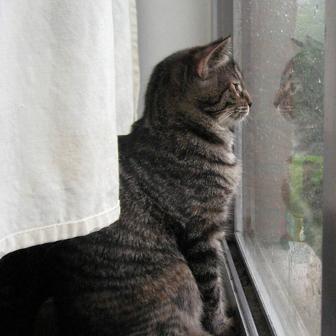}}}&
\raisebox{-0.\height}{\includegraphics[width=\cellwidth cm, height=\cellwidth cm]{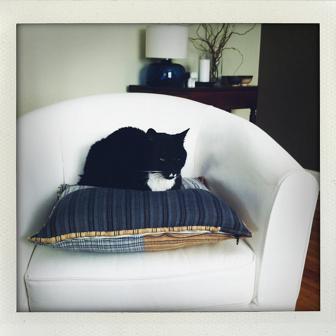}}&
\raisebox{-0.\height}{\highlight{\includegraphics[width=\cellwidth cm, height=\cellwidth cm]{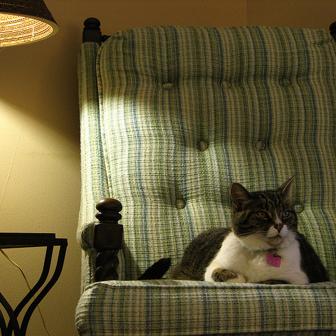}}}
\tabularnewline
\cmidrule{2-11}
 ~  &
\raisebox{-0.\height}{\highlight{\includegraphics[width=\cellwidth cm, height=\cellwidth cm]{samples/resized/3819/000000342078.jpg}}}&
\raisebox{-0.\height}{\highlight{\includegraphics[width=\cellwidth cm, height=\cellwidth cm]{samples/resized/3819/000000070353.jpg}}}&
\raisebox{-0.\height}{\highlight{\includegraphics[width=\cellwidth cm, height=\cellwidth cm]{samples/resized/3819/000000226220.jpg}}}&
\raisebox{-0.\height}{\highlight{\includegraphics[width=\cellwidth cm, height=\cellwidth cm]{samples/resized/3819/000000237419.jpg}}}&
\raisebox{-0.\height}{\highlight{\includegraphics[width=\cellwidth cm, height=\cellwidth cm]{samples/resized/3819/000000132578.jpg}}}&
\raisebox{-0.\height}{\includegraphics[width=\cellwidth cm, height=\cellwidth cm]{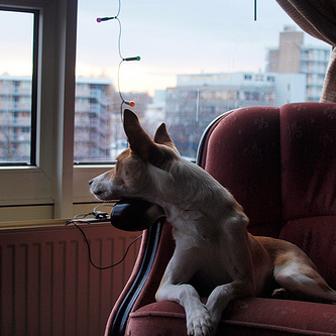}}&
\raisebox{-0.\height}{\highlight{\includegraphics[width=\cellwidth cm, height=\cellwidth cm]{samples/resized/3819/000000070441.jpg}}}&
\raisebox{-0.\height}{\highlight{\includegraphics[width=\cellwidth cm, height=\cellwidth cm]{samples/resized/3819/000000026803.jpg}}}&
\raisebox{-0.\height}{\includegraphics[width=\cellwidth cm, height=\cellwidth cm]{samples/resized/3819/000000125100.jpg}}&
\raisebox{-0.\height}{\includegraphics[width=\cellwidth cm, height=\cellwidth cm]{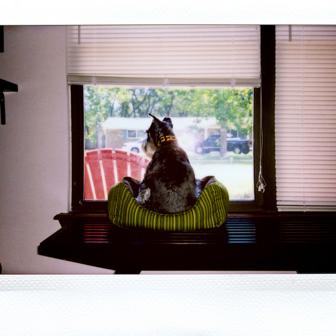}}
\tabularnewline

\bottomrule
\end{tabular}
\end{center}
\vspace{-0.2cm}
\caption{Some examples where our model increases the top-10 ranking similarity in paraphrased text-to-image retrieval. }
    \label{supp:fig:qualitative1}
\end{figure*}

\begin{figure*}[t]
\begin{center}
\setlength{\tabcolsep}{3pt} 
\renewcommand{\arraystretch}{0.0} 
\begin{tabular}{|c|cccccccccc|}
\toprule
 & \multicolumn{10}{|c|}{$q_4$: \textit{``A toy horse \textcolor{red}{sits atop} a red wooden chair.''} (top row)} \tabularnewline
\cmidrule{2-11}
 & \multicolumn{10}{|c|}{$p_4$: \textit{``A toy horse \textcolor{red}{is perched on top of} a red wooden chair.''} (bottom row)} \tabularnewline
\midrule
\multirow{2}{*}{CLIP \cite{radford2021learning}}  &
\raisebox{-0.\height}{\highlight{\includegraphics[width=\cellwidth cm, height=\cellwidth cm]{samples/resized/1494/000000127576.jpg}}}&
\raisebox{-0.\height}{\highlight{\includegraphics[width=\cellwidth cm, height=\cellwidth cm]{samples/resized/1494/000000004979.jpg}}}&
\raisebox{-0.\height}{\highlight{\includegraphics[width=\cellwidth cm, height=\cellwidth cm]{samples/resized/1494/000000468017.jpg}}}&
\raisebox{-0.\height}{\highlight{\includegraphics[width=\cellwidth cm, height=\cellwidth cm]{samples/resized/1494/000000413414.jpg}}}&
\raisebox{-0.\height}{\highlight{\includegraphics[width=\cellwidth cm, height=\cellwidth cm]{samples/resized/1494/000000269932.jpg}}}&
\raisebox{-0.\height}{\highlight{\includegraphics[width=\cellwidth cm, height=\cellwidth cm]{samples/resized/1494/000000276209.jpg}}}&
\raisebox{-0.\height}{\highlight{\includegraphics[width=\cellwidth cm, height=\cellwidth cm]{samples/resized/1494/000000405264.jpg}}}&
\raisebox{-0.\height}{\includegraphics[width=\cellwidth cm, height=\cellwidth cm]{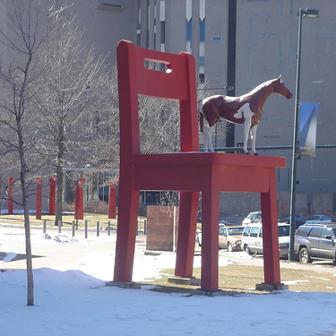}}&
\raisebox{-0.\height}{\highlight{\includegraphics[width=\cellwidth cm, height=\cellwidth cm]{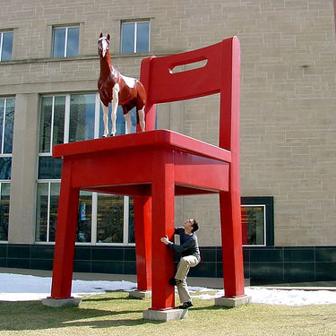}}}&
\raisebox{-0.\height}{\highlight{\includegraphics[width=\cellwidth cm, height=\cellwidth cm]{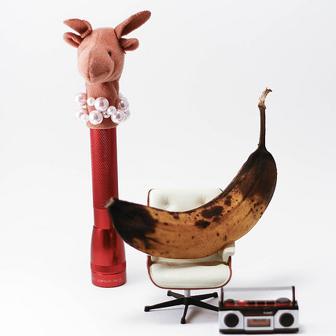}}}
\tabularnewline
\cmidrule{2-11}
 ~  &
\raisebox{-0.\height}{\highlight{\includegraphics[width=\cellwidth cm, height=\cellwidth cm]{samples/resized/1494/000000127576.jpg}}}&
\raisebox{-0.\height}{\highlight{\includegraphics[width=\cellwidth cm, height=\cellwidth cm]{samples/resized/1494/000000468017.jpg}}}&
\raisebox{-0.\height}{\highlight{\includegraphics[width=\cellwidth cm, height=\cellwidth cm]{samples/resized/1494/000000004979.jpg}}}&
\raisebox{-0.\height}{\highlight{\includegraphics[width=\cellwidth cm, height=\cellwidth cm]{samples/resized/1494/000000413414.jpg}}}&
\raisebox{-0.\height}{\highlight{\includegraphics[width=\cellwidth cm, height=\cellwidth cm]{samples/resized/1494/000000405264.jpg}}}&
\raisebox{-0.\height}{\highlight{\includegraphics[width=\cellwidth cm, height=\cellwidth cm]{samples/resized/1494/000000269932.jpg}}}&
\raisebox{-0.\height}{\highlight{\includegraphics[width=\cellwidth cm, height=\cellwidth cm]{samples/resized/1494/000000385016.jpg}}}&
\raisebox{-0.\height}{\highlight{\includegraphics[width=\cellwidth cm, height=\cellwidth cm]{samples/resized/1494/000000163951.jpg}}}&
\raisebox{-0.\height}{\highlight{\includegraphics[width=\cellwidth cm, height=\cellwidth cm]{samples/resized/1494/000000276209.jpg}}}&
\raisebox{-0.\height}{\includegraphics[width=\cellwidth cm, height=\cellwidth cm]{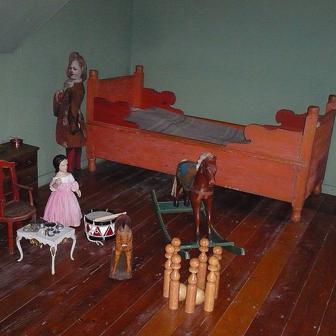}}
\tabularnewline
\midrule
\multirow{2}{*}{Ours}  &
\raisebox{-0.\height}{\highlight{\includegraphics[width=\cellwidth cm, height=\cellwidth cm]{samples/resized/1494/000000004979.jpg}}}&
\raisebox{-0.\height}{\highlight{\includegraphics[width=\cellwidth cm, height=\cellwidth cm]{samples/resized/1494/000000276209.jpg}}}&
\raisebox{-0.\height}{\includegraphics[width=\cellwidth cm, height=\cellwidth cm]{samples/resized/1494/000000269932.jpg}}&
\raisebox{-0.\height}{\highlight{\includegraphics[width=\cellwidth cm, height=\cellwidth cm]{samples/resized/1494/000000407220.jpg}}}&
\raisebox{-0.\height}{\includegraphics[width=\cellwidth cm, height=\cellwidth cm]{samples/resized/1494/000000578072.jpg}}&
\raisebox{-0.\height}{\includegraphics[width=\cellwidth cm, height=\cellwidth cm]{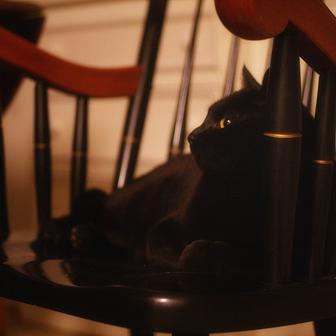}}&
\raisebox{-0.\height}{\includegraphics[width=\cellwidth cm, height=\cellwidth cm]{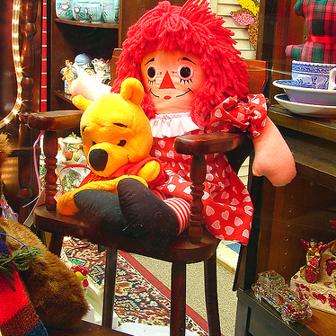}}&
\raisebox{-0.\height}{\highlight{\includegraphics[width=\cellwidth cm, height=\cellwidth cm]{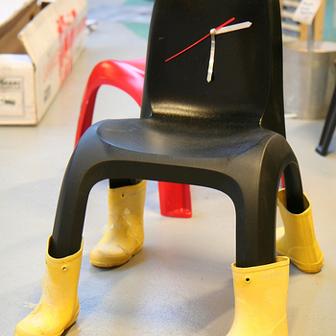}}}&
\raisebox{-0.\height}{\includegraphics[width=\cellwidth cm, height=\cellwidth cm]{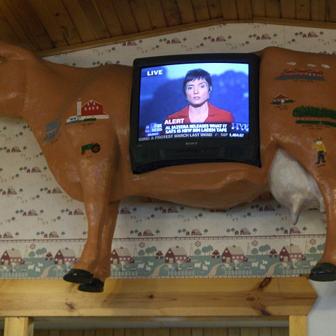}}&
\raisebox{-0.\height}{\highlight{\includegraphics[width=\cellwidth cm, height=\cellwidth cm]{samples/resized/1494/000000417653.jpg}}}
\tabularnewline
\cmidrule{2-11}
 ~  &
\raisebox{-0.\height}{\highlight{\includegraphics[width=\cellwidth cm, height=\cellwidth cm]{samples/resized/1494/000000276209.jpg}}}&
\raisebox{-0.\height}{\highlight{\includegraphics[width=\cellwidth cm, height=\cellwidth cm]{samples/resized/1494/000000004979.jpg}}}&
\raisebox{-0.\height}{\includegraphics[width=\cellwidth cm, height=\cellwidth cm]{samples/resized/1494/000000468017.jpg}}&
\raisebox{-0.\height}{\highlight{\includegraphics[width=\cellwidth cm, height=\cellwidth cm]{samples/resized/1494/000000407220.jpg}}}&
\raisebox{-0.\height}{\includegraphics[width=\cellwidth cm, height=\cellwidth cm]{samples/resized/1494/000000302026.jpg}}&
\raisebox{-0.\height}{\includegraphics[width=\cellwidth cm, height=\cellwidth cm]{samples/resized/1494/000000127576.jpg}}&
\raisebox{-0.\height}{\includegraphics[width=\cellwidth cm, height=\cellwidth cm]{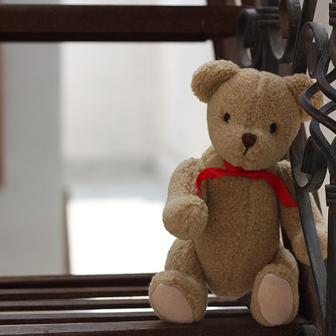}}&
\raisebox{-0.\height}{\highlight{\includegraphics[width=\cellwidth cm, height=\cellwidth cm]{samples/resized/1494/000000538555.jpg}}}&
\raisebox{-0.\height}{\highlight{\includegraphics[width=\cellwidth cm, height=\cellwidth cm]{samples/resized/1494/000000385016.jpg}}}&
\raisebox{-0.\height}{\highlight{\includegraphics[width=\cellwidth cm, height=\cellwidth cm]{samples/resized/1494/000000417653.jpg}}}
\tabularnewline
\bottomrule
\end{tabular}
\end{center}
\vspace{-0.2cm}
\caption{An example where our model fails to increase the top-10 ranking similarity in paraphrased text-to-image retrieval.}
    \label{supp:fig:qualitative2}
\end{figure*}


\clearpage

\end{document}